\pdfoutput=1

\documentclass[11pt]{article}

\usepackage[]{acl}

\usepackage{graphicx}
\usepackage{bbding}
\usepackage{ulem} 
\usepackage{times}
\usepackage{latexsym}
\usepackage{multirow}
\usepackage{pifont}
\usepackage{amsmath}
\usepackage{wrapfig}
\usepackage{booktabs}
\usepackage{hyperref}
\usepackage{tabularx}
\usepackage{csquotes}
\usepackage[tikz]{mdframed}
\usetikzlibrary{shadows}
\usepackage{arydshln}
\usepackage{color, colortbl}
\usepackage{array}
\usepackage[ruled,vlined]{algorithm2e}
\definecolor{commentcolor}{RGB}{110,154,155}   
\newcommand{\PyComment}[1]{\ttfamily\textcolor{commentcolor}{\# #1}}  
\newcommand{\PyCode}[1]{\ttfamily\textcolor{black}{#1}} 
\newcommand{\eg}{\textit{e.g.,}\ }
\newcommand{\ie}{\textit{i.e.,}\ }
\usepackage[T1]{fontenc}

\usepackage[utf8]{inputenc}

\usepackage{microtype}

\definecolor{question}{RGB}{223,223,223}
\definecolor{prompt}{RGB}{230,236,225}
\definecolor{context}{RGB}{236,215,192}
\usepackage{inconsolata}
\usepackage{xcolor}
\definecolor{introred}{RGB}{248,203,173}

\newcommand{\wisdom}{WisdoM\includegraphics[width=0.02\textwidth]{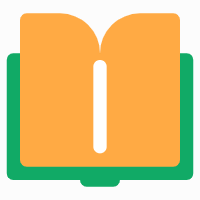}~}
\newcommand{\green}[1]{\textcolor[RGB]{0,176,80}{#1}}
\newcommand{\red}[1]{\textcolor[RGB]{176,0,80}{#1}}

\usepackage{tikz}       %
\usepackage{tcolorbox}
\usepackage{makecell}
\usetikzlibrary{shapes.misc}
\usetikzlibrary{shadows}

\DeclareRobustCommand{\tcgray}[1]{
\begin{tikzpicture}[baseline=(char.base)]
\node(char)[
  draw,fill=black!15,
  shape=rounded rectangle,
  text height=5pt,
  drop shadow={opacity=.5,shadow xshift=0pt,shadow yshift=-1pt},
]
  {\normalfont #1};
\end{tikzpicture}
}
\DeclareRobustCommand{\tcblack}[1]{
\begin{tikzpicture}[baseline=(char.base)]
\node(char)[
  draw,fill=black,
  shape=rounded rectangle,
  text height=5pt,
  drop shadow={opacity=.5,shadow xshift=0pt,shadow yshift=-1pt},
]
  {\color{white}{\normalfont #1}};
\end{tikzpicture}
}
\definecolor{backgroundcolor}{gray}{.9}

\title{WisdoM\includegraphics[width=0.03\textwidth]{fig/wisdom.png}: Improving Multimodal Sentiment Analysis by Fusing Contextual World Knowledge}

\author{Wenbin Wang\textsuperscript{\rm 1}\space\space
Liang Ding\textsuperscript{\rm 2}\space\space
Li Shen\textsuperscript{\rm 3}\space\space
Yong Luo\textsuperscript{\rm 1}\space\space
Han Hu\textsuperscript{\rm4}\space\space
Dacheng Tao\textsuperscript{\rm 2}\\
    \textsuperscript{\rm 1}Wuhan University\space\space
    \textsuperscript{\rm 2}The University of Sydney\\ 
    \textsuperscript{\rm 3}JD Explore Academy\space\space
    \textsuperscript{\rm 4}Beijing Institute of Technology\\
    {\tt\small \{wangwenbin97, luoyong\}@whu.edu.cn},\space\space
    {\tt\small liangding.liam@gmail.com}
}

\begin{document}
\maketitle
\begin{abstract}
Sentiment analysis is rapidly advancing by utilizing various data modalities (\eg text, image). However, most previous works relied on superficial information, neglecting the incorporation of contextual world knowledge (\eg background information derived from but beyond the given image and text pairs) and thereby restricting their ability to achieve better multimodal sentiment analysis (MSA).
In this paper, we proposed a plug-in framework named \wisdom, to leverage the contextual world knowledge induced from the large vision-language models (LVLMs) for enhanced MSA. WisdoM utilizes LVLMs to comprehensively analyze both images and corresponding texts, simultaneously generating pertinent \textit{context}. To reduce the noise in the context, we also introduce a training-free contextual fusion mechanism. Experiments across diverse granularities of MSA tasks consistently demonstrate that our approach has substantial improvements (brings an average +1.96\% F1 score among five advanced methods) over several state-of-the-art methods. The code will be released.
\end{abstract}

\section{Introduction}

Sentiment analysis (SA,~\citealp[]{MEDHAT20141093,wankhade2022survey}) task focuses on identifying human sentiment polarity~\cite{das2023multimodal}. 
With the development of the Internet, the way people express their sentiments is not limited to text, but also includes multimodal data (\eg images). 
Detecting the sentiments of such information is challenging because of their short, informal nature, but utilizing the paired images can offer valuable insights. Therefore, how to make accurate sentiment classification by effectively marring modalities is at the core of multimodal sentiment analysis (MSA,~\citealp[]{majumder2018multimodal,wang2020transmodality}).

Recent studies that improve MSA by carefully designing fusion strategies can be categorized into two types: 1) for aspect-level MSA, existing methods~\cite{ju2021joint,ling2022vision,yang2022cross,zhou-etal-2023-aom} estimate whether the image contains conducive information by learning the relationship between images and text, and locally fuse the aspect-aware images; 2) for sentence-level MSA, researchers primarily focus on the global fusion at feature-level and decision-level~\cite{hazarika2020misa,yu2021learning,han2021improving,guo2022dynamically}.
\begin{figure}[t]
    \centering
    \includegraphics[width=1.0\linewidth]{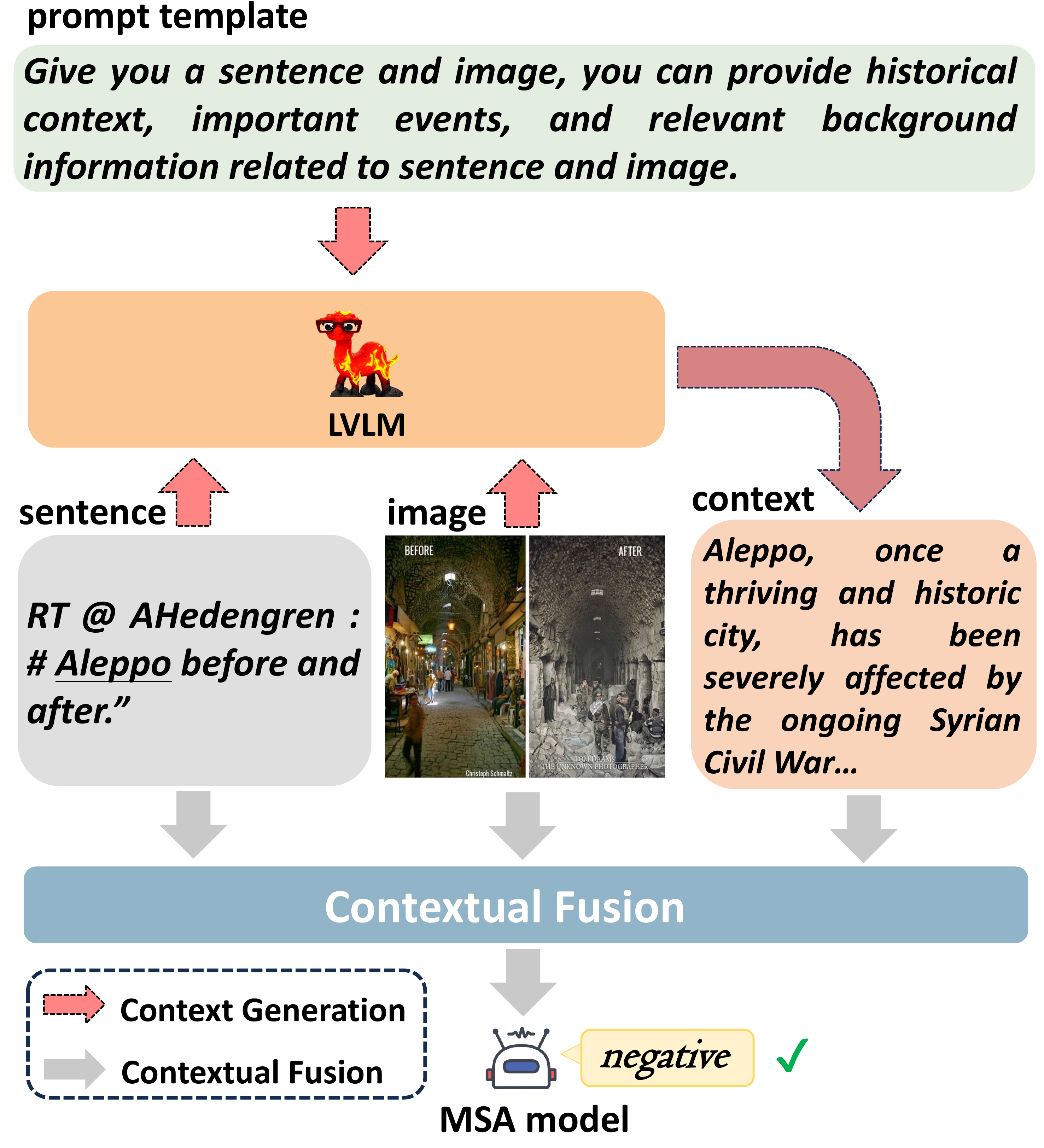}
    \caption{\textbf{The simple schematic of our method}. The sentiment polarity of \underline{Aleppo} is negative, which is hard to directly predict by existing methods while our \wisdom predicts correctly via incorporating \textit{context} generated by the world knowledge-rich LVLMs.}
    \label{fig:motivation}
\end{figure}
Despite their empirical success, the above studies only consider the \textit{superficial information}\footnote{only reflects the surface or literal information without considering their deep (\eg historical and cultural) meaning.} between image and text (See the case ``\textit{Aleppo}'' in Fig.~\ref{fig:motivation}), and sometimes it is difficult to predict the true polarity without their background world knowledge (``\textit{Aleppo has been severely affected by the ongoing Syrian Civil War...}'' induced from the large vision-language models.).
This raises the following question:
\begin{mdframed}[backgroundcolor=gray!40,shadow=true,roundcorner=8pt]
\centering{\textbf{Could world knowledge boost MSA?}}
\end{mdframed}
Take the test case in Fig.~\ref{fig:motivation} as an instance, given a comparative image at different periods alongside a sentence, it is required to answer the question: \textit{\colorbox{question}{What's the sentiment polarity of ``Aleppo''?}}. We employ the current state-of-the-art (SOTA) MSA model~\cite{zhou-etal-2023-aom} as the backbone. As expected, even the SOTA model gives the wrong prediction (\ie ``neutral'' rather than the groundtruth-- ``negative'') because of lacking the deeper knowledge of ``Aleppo'' (which is a city in Syria, in conjunction with the image, we might infer that the difference between the before and after of this city is caused by the Syrian war). Therefore, employing world knowledge is essential.

Correspondingly, we propose a plug-and-play framework to utilize the contextual world knowledge (simply induced from the large vision-language models) to complement the existing text-image pair with only superficial information, namely \wisdom. In particular, our WisdoM follows a three stages process:
\tcblack{\small{1}}Prompt Templates Generation, \tcblack{\small{2}}Context Generation, and \tcblack{\small{3}}Contextual Fusion. In \textbf{stage 1}, we ask language models (\eg ChatGPT\footnote{\url{https://chat.openai.com}}) to generate prompt templates (See the Prompt Template ``\textit{\colorbox{prompt}{Give you a sentence and image, you can ...}}'' in Fig.~\ref{fig:motivation}) which are used to construct instructions for \textbf{stage 2}. Then, we employ the advanced large vision-language model (LVLM, \eg LLaVA~\citealp[]{liu2023llava} in \textbf{stage 2} to generate the contextual information (See the Context ``\textit{\colorbox{context}{Aleppo, ... ongoing Syrian Civil War...}}'' in Fig.~\ref{fig:motivation}) based on the provided image and sentence. Note that, we refer to this contextual information as \textit{context}. Due to the noisy nature of the derived \textit{context}, we further introduce a training-free Contextual Fusion mechanism, in \textbf{stage 3}, to wisely incorporate the \textit{context} for hard samples. 

We validated our WisdoM on several benchmarks including Twitter2015, Twitter2017~\cite{yu2019adapting} and MSED~\cite{jia2022beyond} over several models: LLaVA-v1.5~\cite{liu2023improvedllava}, MMICL~\cite{zhao2023mmicl}, mPLUG-Owl2~\cite{ye2023mplugowl2}, AoM~\cite{zhou-etal-2023-aom}, ALMT~\cite{zhang2023learning}. Experiments demonstrate the effectiveness and universality of our approach, and extensive analyses provide insights into when and how our method works. Our main \textbf{contributions} are:
\begin{itemize}
    \item We propose a plug-in framework WisdoM, leveraging the LVLM to generate explicit contextual world knowledge, to enhance the multimodal sentiment analysis ability.
    \item To achieve wise knowledge fusion, we introduce a novel contextual fusion mechanism to mitigate the impact of noise in the \textit{context}.
    \item Experiments on three MSA benchmarks upon several advanced LVLMs, show that WisdoM brings consistent and significant improvements (up to +6.3\% F1 score).
\end{itemize}

\section{Related Work}
\subsection{Multimodal Sentiment Analysis}
Multimodal Sentiment Analysis (MSA), diverging from conventional text-based approaches~\cite{hussein2018survey}, incorporates diverse modalities (\eg image, speech) to enhance sentiment classification accuracy~\cite{soleymani2017survey}. Numerous advanced models have been proposed, covering different levels of granularity, such as sentence and aspect:
\paragraph{Sentence-Level MSA.}
\citet{wang2014microblog} integrate images and text for microblog analysis. \citet{you2015joint} employ deep neural networks for textual and visual sentiment analysis. \citet{zhao2019image} explore image-text correlations in movie reviews. \citet{li2020convolutional} propose a ConvTransformer, blending Transformer~\cite{vaswani2017attention} and CNN technologies for sentiment analysis. \citet{das2022multi} propose a multi-stage multimodal method for the Assamese language, leveraging both text and images. \citet{zhang2023learning} present an advanced model, ALMT, enhancing multimodal analysis with a focus on language-guided features to handle irrelevant or conflicting data across different modalities.
\paragraph{Aspect-Level MSA.}
\citet{yu2019adapting} introduce TomBERT, leveraging two multimodal tweet datasets with target annotations, and proposed an aspect-oriented multimodal BERT model. \citet{khan2021exploiting} propose a two-stream model employing an object-aware transformer for image translation in the input space, followed by a single-pass non-autoregressive generation approach~\cite{wu2020slotrefine,ding2021understanding}. \citet{zhou-etal-2023-aom} introduce an aspect-oriented network, AoM, designed to reduce visual and textual distractions arising from intricate image-text interactions.

Although existing approaches relying on sophisticated fusion techniques have achieved remarkable performance in MSA, their limitation lies in relying on superficial information for fusion, without incorporating contextual world knowledge. 

\subsection{Large Vision-Language Models}

Large Vision-Language Models (LVLMs) are becoming a fundamental tool for solving general tasks~\cite{li2023seed, fu2023mme,liu2023llava,zhao2023mmicl,ye2023mplugowl2,Dai2023InstructBLIPTG,zhu2023minigpt,alayrac2022flamingo,chen2023shikra}. \citet{liu2023improvedllava} introduce LLaVA, which connects the CLIP ViT-L/14 visual encoder~\cite{dosovitskiy2020image} with the large language model Vicuna~\cite{chiang2023vicuna} through a simple projection matrix, utilizing a two-stage instruction-tuning method. \citet{zhao2023mmicl} propose MMICL to tackle the complexity of multi-modal prompts, focusing on improving LVLMs from both model and data viewpoints. \citet{ye2023mplugowl2} propose mPUG-Owl2, a multi-modal language model designed for enhanced collaboration between modalities. It features a modular network where the language decoder acts as a universal interface for different modalities.

Here, we leverage LVLMs to enhance multimodal sentiment analysis by generating relevant world knowledge. Additionally, we introduce a new training-free module named Contextual Fusion, designed to minimize noise in the context.

\subsection{Retrieval-Augmented Generation}
Retrieval-Augmented Generation (RAG) enhances Language Models by incorporating retrieved text, significantly improving performance in knowledge-based tasks, applicable in both fine-tuned and off-the-shelf scenarios~\cite{gao2023retrieval,gupta2024rag}.
Traditional RAG~\cite{lewis2020retrieval}, also known as Naive RAG, incorporates retrieval content to aid generation but faces key challenges: 1) varying retrieval quality, 2) generation of responses prone to inaccuracies, and 3) difficulties in coherently integrating retrieved-context with current tasks. To overcome the limitations of Naive RAG, advanced methods introduce more contextually rich information during inference. The DSP framework~\cite{khattab2022demonstrate} facilitates an intricate exchange between frozen LMs and retrieval models, improving context richness, while PKG~\cite{luo2023augmented} allows LLMs to retrieve relevant information for complex tasks without altering their parameters. The working mechanism of WisdoM is similar to RAG, but \textbf{different} at the following aspects: \ding{172} WisdoM utilizes LVLM to generate world knowledge to provide coherent and accurate context rather than retrieval, \ding{173} WisdoM incorporates a contextual fusion mechanism to diminish noise within the context. For additional experimental analysis and discussion, please refer to \S~\ref{sec:effect_context}.

\begin{figure}
    \centering
    \includegraphics[width=0.9\linewidth]{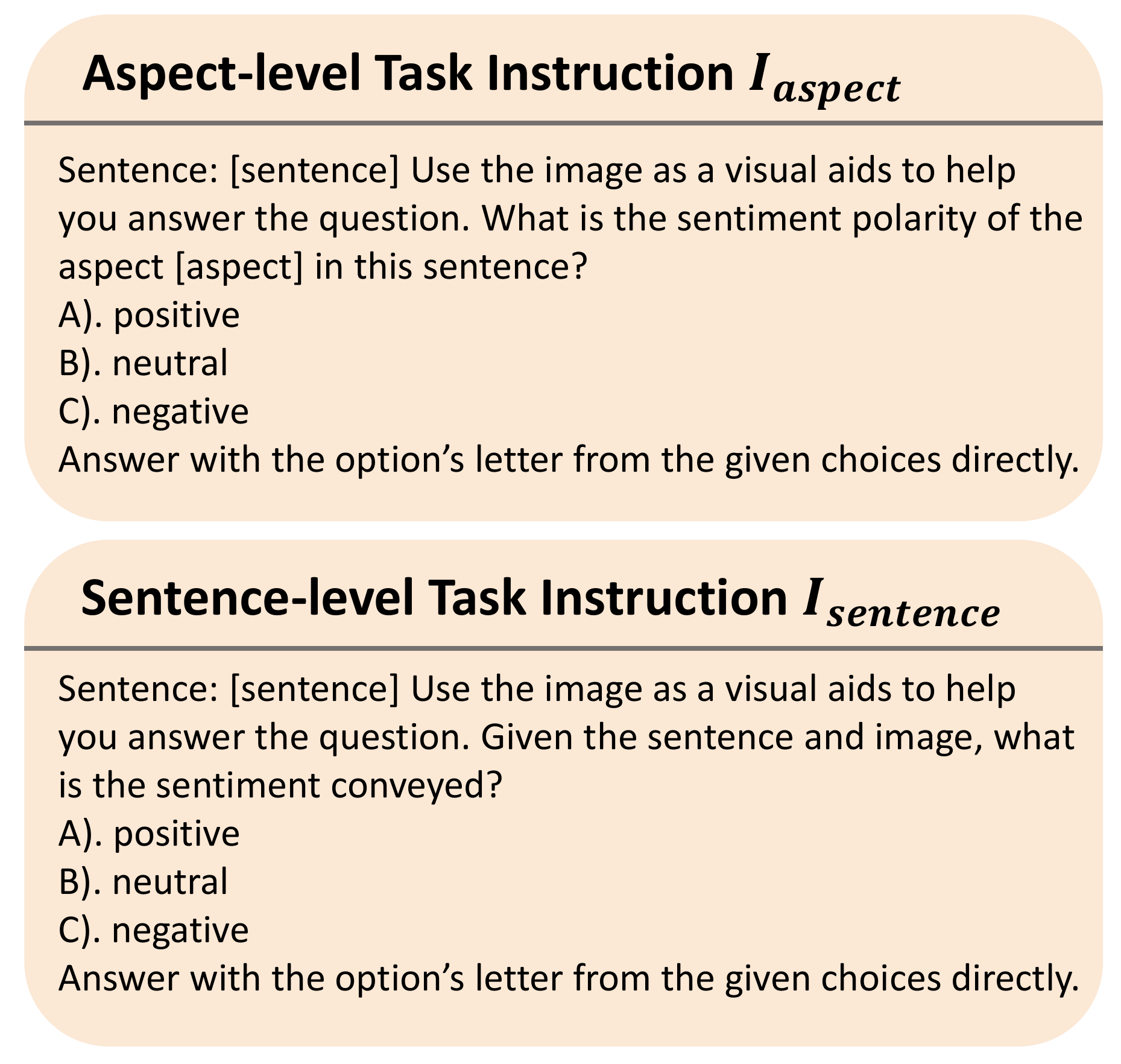}
    \caption{\textbf{Template of task instruction}.}
    \label{fig:task_instruction}
\end{figure}
\begin{figure*}[t]
    \centering
    \includegraphics[width=1.\linewidth]{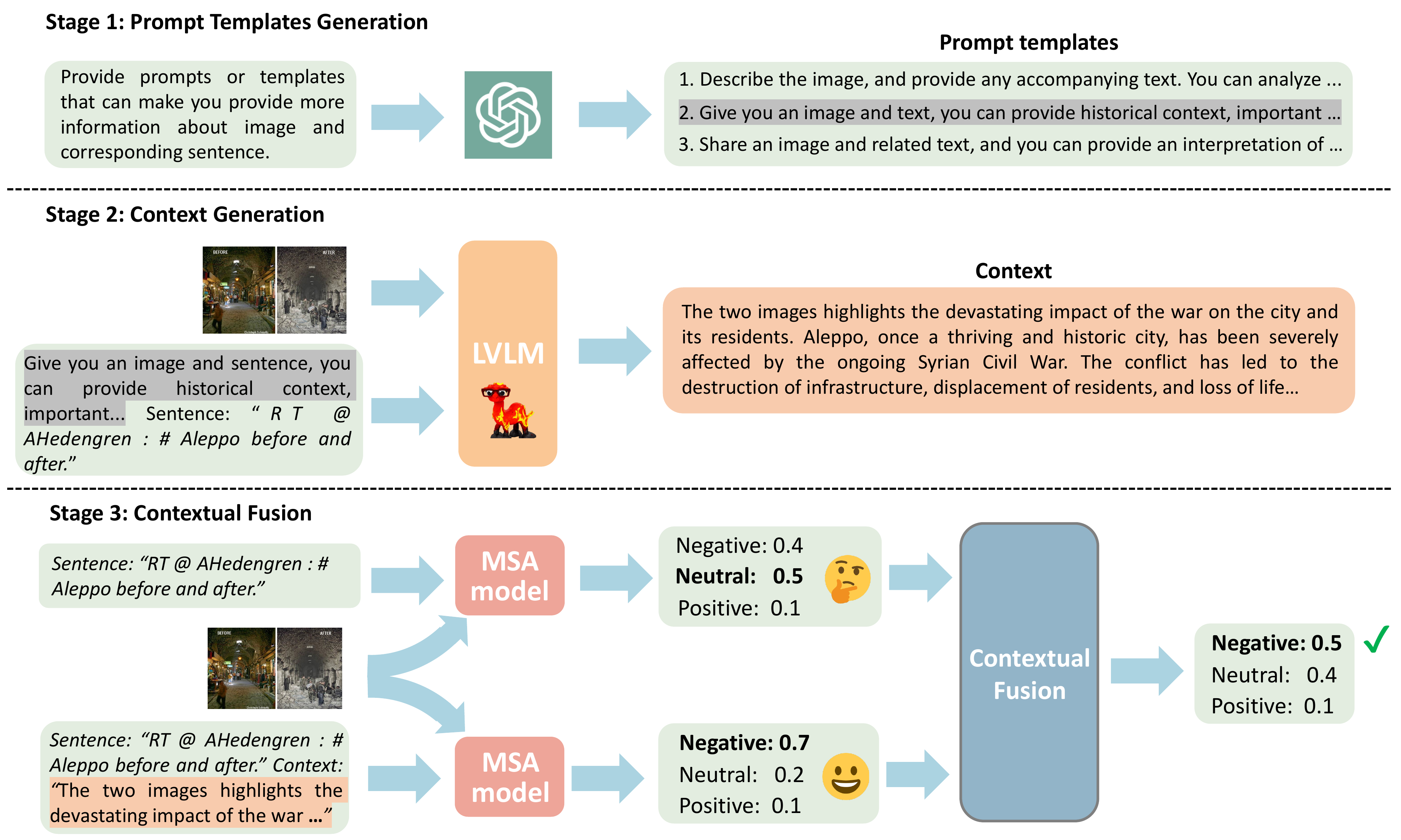}
    \caption{\textbf{Detailed illustration of our proposed schema} \wisdom with a running example. \tcgray{\small{1}} Using ChatGPT to provide prompt templates. \tcgray{\small{2}} We then prompt LVLMs to generate \textit{context} using the prompt templates with image and sentence. \tcgray{\small{3}} A training-free mechanism Contextual Fusion mitigates the noise in the \textit{context}.}
    \label{fig:architecture}
\end{figure*}
\section{Preliminary}
We first describe the notation of the MSA, then review two typical frameworks for modelling the MSA tasks, where we experiment with our schema upon them: task-specific framework~\cite{zhou-etal-2023-aom,zhang2023learning} and general-purpose framework~\cite{liu2023improvedllava,ye2023mplugowl2}.
\paragraph{Notation.}
Let $\mathcal{M}$ be a set of multimodal samples. Each sample $m_i \in \mathcal{M}$ consists of a sentence $s_i$ and image $v_i$. For \textit{aspect}-level MSA tasks, there are several aspects $a_i$ which is a subsequence of $s_i$, \ie $a_i \in s_i$. We denote $f(\cdot)$ as the sentiment classifier. The output of $f(\cdot)$ is the sentiment polarity $y_i \in $ \{negative, neutral, positive\}, with corresponding predicted probability denoted as $P_i=\{p^{neg}_{i}, p^{neu}_i, p^{pos}_i\}$.

\paragraph{Task-Specific Framework.}
For \textit{aspect}-level MSA tasks, the goal is to predict the sentiment polarity $y_i$ and probability $P_i$ for the specific aspect $a_i$ conditioned on the $(v_i,s_i)$, \ie $(y_i,P_i) = f(v_i,s_i,a_i)$. For \textit{sentence}-level MSA tasks, the $y_i$ and $P_i$ are predicted by sentence $s_i$ alongside the image $v_i$, \ie $(y_i,P_i) = f(v_i,s_i)$. 
\paragraph{General-Purpose Framework.}
To verify that our WisdoM works well on arbitrary architectures, we also apply WisdoM to general-purpose LVLMs. We follow~\citet{wang2023chatgpt} to construct the task instructions $I_{aspect}$ and $I_{sentence}$ for each task to elicit its ability to the corresponding task. The task instructions are presented as single-choice questions with well-formatted options (Shown in Fig.~\ref{fig:task_instruction}). The LVLMs can be seen as a sentiment classifier $f(\cdot)$. Therefore, the \textit{aspect}-level task and \textit{sentence}-level task can be formulated as $(y_i,P_i) = f(I_{aspect}, v_i, s_i, a_i)$ and $(y_i,P_i) = f(I_{sentence}, v_i, s_i)$ respectively.

\section{Methodology}
\paragraph{Overview.}
Fig.~\ref{fig:architecture} illustrates the overview of our method following three stages. In the \tcblack{\small{1}}Prompt Templates Generation, we use large language models, particularly ChatGPT, to provide prompt templates. These prompt templates are fed into the LVLM with sentence $s_i$ and image $v_i$ to generate \textit{context}, also called the \tcblack{\small{2}}Context Generation. During \tcblack{\small{3}}Contextual Fusion, we first compute the confidence, determining if the sample is uncertain (referred to as a hard sample). For hard samples, we fuse the predicted probability $P_i$ with $\hat{P}_i$ which is obtained by incorporating \textit{context}. Otherwise, we use $P_i$ as the final prediction.


\subsection{Stage 1: Prompt Templates Generation}
The main purpose of this stage is to design the prompt templates used to generate the \textit{context}, so that the LVLM can better understand our intention and thus provide a more comprehensive contextual world knowledge. Inspired by~\cite{jiao2023chatgpt,zhong2023can}, we ask ChatGPT to provide the appropriate prompt templates. The prompt templates provided by ChatGPT consider world knowledge of different perspectives, including history, social and cultural, \textit{etc.}  We insert a ``Sentence: [x]'' at the end of the prompt template to place the input sentence $s_i$. The example of prompt templates are shown in Appendix~\ref{app:prompt_templates}. 


\begin{table*}[]
\centering
\begin{tabular}{lllll}
\toprule
       \multirow{2}{*}{\bf Method}            & \multicolumn{2}{c}{\bf Twitter2015} & \multicolumn{2}{c}{\bf Twitter2017} \\ \cmidrule(lr){2-3} \cmidrule(lr){4-5} 
          & \textbf{\textit{Acc.}}            & \textbf{\textit{F1}}         & \textbf{\textit{Acc.}}       & \textbf{\textit{F1}}             \\ \hline
ESAFN~\cite{yu2019entity}            & 73.4           & 67.4           & 67.8       & 64.2               \\
TomBERT~\cite{yu2019adapting}         & 77.2           & 71.8           & 70.3       & 68.0               \\
CapTrBERT~\cite{khan2021exploiting}        & 77.9           & 73.9           & 72.3       & 70.2               \\
JML~\cite{ju2021joint}             & 78.7           & -              & 72.7       & -                  \\
VLP-MABSA~\cite{ling2022vision}       & 78.6           & 73.8           & 73.8       & 71.8               \\
CMMT~\cite{yang2022cross}          & 77.9           & -              & 73.8       & -                  \\
\hline
MMICL~\cite{zhao2023mmicl}*            & 76.0        & 72.7          & 74.1      & 74.0              \\ 
\quad \textbf{-w/ WisdoM}     & 77.3\ {\small (\green{+1.3})}          & 74.2\ {\small (\green{+1.5})}          & 75.7\ {\small (\green{+1.6})}      & 75.7\ {\small (\green{+1.1})}              \\ \hdashline
LLaVA-v1.5~\cite{liu2023improvedllava}*      &      77.9          &         74.3       &     74.6       &       74.3             \\ 
\quad \textbf{-w/ WisdoM}  &    78.9\ {\small (\green{+1.0})}        &      75.6\ {\small (\green{+1.3})}          &     75.6\ {\small (\green{+1.0})}       &         75.3\ {\small (\green{+1.0})}           \\ \hdashline
mPLUG-Owl2~\cite{ye2023mplugowl2}*  & 76.8 & 72.3 & 74.2 & 73.0 \\ 
\quad \textbf{-w/ WisdoM}  &  77.3\ {\small (\green{+0.5})} & 73.4\ {\small (\green{+1.1})} & 74.5\ {\small (\green{+0.3})} & 73.7\ {\small (\green{+0.7})} \\ \hdashline
AoM~\cite{zhou-etal-2023-aom}*            & 80.0           & 75.2           & 75.9       & 74.5                 \\
\quad \textbf{-w/ WisdoM}      & \textbf{81.5}\ {\small (\green{+1.5})} & \textbf{78.1}\ {\small (\green{+2.9})} & \textbf{77.6}\ {\small (\green{+1.7})}      & \textbf{76.8}\ {\small (\green{+2.3})}     \\
\bottomrule
\end{tabular}
\caption{\textbf{Comparison of our method (upon several advanced models) with existing works} on Twitter2015 and 2017 benchmarks. The highest results are highlighted in bold, and * indicates the reproduced results.}
\label{table:msc_result}
\end{table*}
\subsection{Stage 2: Context Generation}
In the context generation stage, prompt templates in \tcblack{\small{1}} are used to generate \textit{context} that explicitly incorporates world knowledge based on given the image $v_i$ and sentence $s_i$ by LVLMs~\cite{liu2023llava,ye2023mplugowl2}. 
Specifically, we construct instruction by replacing the ``[x]'' in the prompt template with the sentence $s_i$. In addition, different LVLMs require a special token to indicate where the image $v_i$ is inserted. Taking LLaVA~\cite{liu2023llava} as an example, we insert a special token ``<image>'' at the beginning of the instruction.

\subsection{Stage 3: Contextual Fusion}
We use the \textit{sentence}-level task as an example. After obtaining the \textit{context}, we can intuitively use predicted sentiment polarity $\hat{y}_i$ obtained by incorporating \textit{context}, \ie $(\hat{y}_i, \hat{P}_i)=f(v_i,s_i,context)$. However, the \textit{context} may contain irrelevant information that could disturb performance. Therefore, we first determine the hard samples and then fuse the $P_i$ and $\hat{P}_i$ in the hard sample. 
\paragraph{Determining the Hard Samples.}
Inspired by~\citet{zhang2021cola}, we found that the ambiguous hard sample is commonly found around boundary areas of the sentiment polarity (\eg the boundary areas of negative and neutral). Therefore, given a sample $m_i$, we only consider the difference $\delta_i$ between the highest and the second highest probabilities to determine whether it is a hard sample:
\begin{equation}
    \delta_i = 2 \times max(P_i) + min(P_i) - 1.
\end{equation}
Then, we denote uncertain threshold $\alpha$ to select samples that not exceed $\alpha$ as hard, \ie$\mathcal{V}_{hard}=\{m_i|\delta_i \leq \alpha\}$, where we leave $\alpha=0.3$ as default.
\paragraph{Fusion with Context.}
Inspired by~\cite{li2022contrastive,o2023contrastive}, we take the convex combinations of $P_i$ and $\hat{P}_i$ to obtain the final prediction $\tilde{P}_i$ for hard sample $m_i \in \mathcal{V}_{hard}$:
\begin{equation}
    \tilde{P}_i = P_i + \beta \cdot (\hat{P}_i - P_i),
\end{equation}
where $\beta$ is an interpolation coefficient. Intuitively, $(\hat{P}_i - P_i)$ represents the information incorporating by \textit{context}. $\beta$ is used to control the proportion of information introduced in \textit{context}. When $\beta \rightarrow 0$, the effect brought by \textit{context} is completely ignored and vice versa. Note that, we use $(y_i,P_i)$ as the final prediction when $m_i \notin \mathcal{V}_{hard}$. We study the impact of $\alpha$ and $\beta$ in Appendix~\ref{section:hyperparameter}.

\section{Experiments}
In this section, we apply WisdoM to \textit{aspect}-level and \textit{sentence}-level MSA tasks to verify its effectiveness and conduct extensive analysis to better understand the proposed method.

\begin{table*}[t]
\centering
\begin{tabular}{llll}
\toprule
\multirow{2}{*}{\bf Method}  & \multicolumn{3}{c}{\bf MSED}                         \\ \cline{2-4} 
                                           & \multicolumn{1}{l}{\textbf{\textit{Precision}}}            & \multicolumn{1}{l}{\textbf{\textit{Recall}}}              & \multicolumn{1}{l}{\textbf{\textit{F1}}}         \\ \hline
DCNN+AlexNet~\cite{jia2022beyond}                        & 71.02          & 70.09          & 70.31          \\
DCNN+ResNet~\cite{jia2022beyond}                             & 74.73          & 74.73          & 74.64          \\
BiLSTM+AlexNet~\cite{jia2022beyond}                        & 78.73          & 79.22          & 78.89          \\
BERT+AlexNet~\cite{jia2022beyond}                          & 83.22          & 83.11          & 83.16          \\
Multimodal Transformer~\cite{jia2022beyond}                 & 83.56          & 83.45          & 83.50          \\
\hline
ALMT~\cite{zhang2023learning}*                                    & 83.73          & 83.98          & 83.73          \\
\quad \textbf{-w/ WisdoM}                                       & 89.92\ {\small (\green{+6.19})}          & 90.14\ {\small (\green{+6.16})}          & 90.01\ {\small (\green{+6.28})}          \\ \hdashline
MMICL~\cite{zhao2023mmicl}*                                 & 86.24          & 86.55          & 86.17          \\
\quad \textbf{-w/ WisdoM}                                 & 89.92\ {\small (\green{+3.68})}          & 88.24\ {\small (\green{+1.69})}          & 88.85\ {\small (\green{+2.68})}          \\ \hdashline
mPLUG-Owl2~\cite{ye2023mplugowl2}*                          & 87.75          & 88.28          & 87.98          \\ 
\quad \textbf{-w/ WisdoM}                             & 88.72\ {\small (\green{+0.97})}          & 89.35\ {\small (\green{+1.07})}          & 88.95\ {\small (\green{+0.97})}          \\ \hdashline
LLaVA-v1.5~\cite{liu2023improvedllava}*                              & 88.98          & 88.77          & 88.75          \\
\quad \textbf{-w/ WisdoM}                          & \textbf{90.58}\ {\small (\green{+1.60})} & \textbf{90.41}\ {\small (\green{+1.64})} & \textbf{90.48}\ {\small (\green{+1.73})} \\
\bottomrule
\end{tabular}
\caption{\textbf{Performance of applying our WisdoM to advanced models} on MSED benchmark, with reference results from existing works. The best results are bolded, and the * denotes the reproduced results.}
\label{table:msed_result}
\end{table*}
\subsection{Experimental Settings}
\paragraph{Datasets.}
For aspect-level tasks, our two benchmark datasets are Twitter2015 and Twitter2017~\cite{yu2019adapting}. 
Twitter2015 and Twitter2017 are comprised of multimodal tweets, where each tweet incorporates textual content, an accompanying image, aspects contained within the tweet, and the sentiment associated with each aspect. Each aspect is assigned a label from the predefined set \{negative, neutral, positive\}.
For sentence-level tasks, we evaluate our WisdoM on a multimodal and multi-task dataset MSED~\cite{jia2022beyond}, containing 9,190 text-image pairs.
\paragraph{Models.}
\label{sec:models}
To demonstrate WisdoM generalizes across architectures and sizes, we experimented on MMICL (14B)~\cite{zhao2023mmicl}, LLaVA-v1.5 (13B)~\cite{liu2023improvedllava}, mPLUG-Owl2 (8.2B)~\cite{ye2023mplugowl2}, AoM (105M)~\cite{zhou-etal-2023-aom}, ALMT (112.5M)~\cite{zhang2023learning}. 

\paragraph{Evaluation Metrics.}
For aspect-level tasks, we use Accuracy (\textit{\textbf{Acc.}}) and macro-F1 (\textit{\textbf{F1}}) following previous studies~\cite{khan2021exploiting,ju2021joint,ling2022vision,zhou-etal-2023-aom}. For the sentence-level task, we adopt precision, recall, and macro-F1 (\textit{\textbf{F1}}) as evaluation metrics.
\subsection{Main Results}
\paragraph{Results of Aspect-Level MSA Task.}
We compare against advanced aspect-level MSA methods on Twitter2015 and Twitter2017, and report the results on Table~\ref{table:msc_result}. We show that our WisdoM achieves consistent and significant improvement on four models across two datasets. The WisdoM brings max 2.9\% and 2.3\% F1-gains on Twitter2015 and Twitter2017 respectively, showing that our method has a clear advantage. 

\paragraph{Results of Sentence-Level MSA Task.}
As shown in Table~\ref{table:msed_result}, notably, our WisdoM upon LLaVA-v1.5 achieves the \textbf{new SOTA} F1 score: 90.48\%, outperforming LLaVA-v1.5 (88.75\%), mPLUG-Owl2 (87.98\%), MMICL (86.17\%) and ALMT (83.73\%), consistently. The most significant improvement is achieved on ALMT (112.5M), where we bring an encouragingly 6.28\% F1 gain, suggesting that contextual world knowledge is particularly crucial for small models in MSA.

\begin{table}[t]
\centering
\resizebox{1.\linewidth}{!}{
\begin{tabular}{llll}
\toprule
                   \multirow{2}{*}{\textbf{Method}}           & \multicolumn{2}{c}{\textbf{Twitter2017}} & \multicolumn{1}{c}{\textbf{MSED}} \\ \cmidrule(lr){2-3} \cmidrule(lr){4-4} & \multicolumn{1}{c}{\textbf{\textit{Acc.}}} & \multicolumn{1}{c}{\textbf{\textit{F1}}}  & \multicolumn{1}{c}{\textbf{\textit{F1}}} \\ \hline
\multicolumn{4}{c}{MMICL} \\ \hline
Baseline                    &  74.1   & 74.1       & 86.2                \\ \hdashline
\quad + \textit{context}    & 72.6 {\small(\red{-1.5})} & 74.4  {\small(\green{+0.3})}    & 86.8  {\small(\green{+0.6})}            \\
\quad + CF      & \textbf{75.7} {\small(\green{+1.6})} & \textbf{75.7}  {\small(\green{+1.6})}       & \textbf{88.9}  {\small(\green{+2.7})}      \\ \hline
\multicolumn{4}{c}{LLaVA-v1.5} \\ \hline
Baseline    & 74.6   &  74.3  & 88.8  \\ \hdashline
\quad + \textit{context}   & 73.7 {\small(\red{-0.9})} & 73.5 {\small(\red{-0.8})}   & 87.8  {\small(\red{-1.0})}             \\
\quad + CF      & \textbf{75.6}{\small(\green{+1.0})} & \textbf{75.3}  {\small(\green{+1.0})}        & \textbf{90.5}  {\small(\green{+1.7})}    \\ 
\bottomrule
\end{tabular}
}
\caption{\textbf{Ablation study of \textit{context} and its wise fusion module.} ``CF'' denotes our Contextual Fusion. We first only incorporate ``\textit{context}'' and subsequently introduce the ``contextual fusion'' module.}
\label{table:ablation}
\end{table}

\begin{figure*}[h]
\vspace{-10pt}
    \centering
    \includegraphics[width=0.9\linewidth]{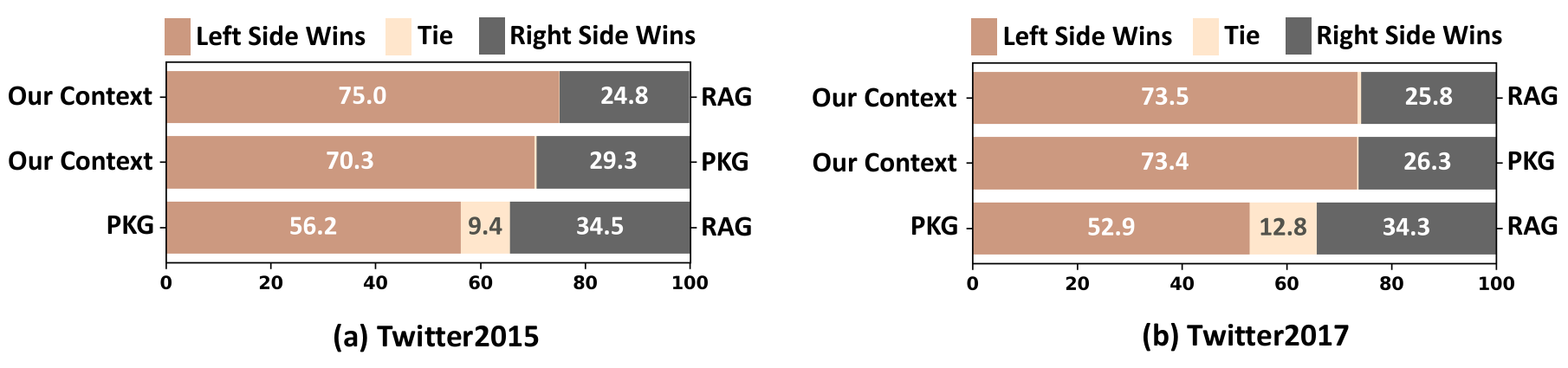}
    \caption{\textbf{Comparative winning rates of Our Context \textit{v.s.} RAG-based methods on Twitter2015 and Twitter2017 benchmarks.} We can see that our contexts are better than the knowledge provided by RAG and PKG.}
    \label{fig:rag_ana}
    \vspace{-6pt}
\end{figure*}

\begin{table}[tb]
    \centering
    \resizebox{1.\linewidth}{!}{
        \begin{tabular}{lcccc}
    \toprule
       \multirow{2}{*}{\textbf{Method}}  &  \multicolumn{2}{c}{\textbf{Twitter2015}} & \multicolumn{2}{c}{\textbf{Twitter2017}}  \\ \cmidrule(lr){2-3} \cmidrule(lr){4-5}
         & \textbf{\textit{Acc.}} & \textbf{\textit{F1}} & \textbf{\textit{Acc.}} & \textbf{\textit{F1}} \\ \hline
         RAG &  75.1  &  71.0  & 71.9 & 70.7 \\
         PKG &  76.2  &  72.4  & 72.8  & 71.5 \\
         \textbf{Our Context} &    \textbf{76.3}        &      \textbf{72.7}           &     \textbf{73.7}       &         \textbf{73.5} \\
    \bottomrule
    \end{tabular}
    }
    \caption{\textbf{Comparative results of \textit{context} generated by RAG, PKG, and our stage 2}. We incorporate the \textit{contexts} on LLaVA-v1.5 directly.}
    \vspace{-10pt}
    \label{tab:exp_rag}
\end{table}

\subsection{Ablation Study}
\label{sec:abaltion}
To better understand the role of each module in our method WisdoM, Table~\ref{table:ablation} presents the ablation results of the gradual addition of different components. Compared with the baselines (MMICL and LLaVA-v1.5), only adding \textit{context} results in a slight performance degradation (-0.23\% average F1 score), while with the help of our proposed Context Fusion mechanism, we achieve a consistent and significant improvement (+1.75\% average F1 score). Through (error) case studies in Appendix~\ref{section:case_study}, we found that containing irrelevant information in the original context leads to bad performance, showing the necessity of further context fusion mechanism.

\paragraph{Analyzing Effects of \textit{Context}.}
\label{sec:effect_context}
To further analyse the effect of \textit{context}, we compare our \textit{context} generated in \textbf{stage 2} with the document retrieved by RAG~\cite{lewis2020retrieval} and knowledge generated by PKG~\cite{luo2023augmented}, collectively termed as ``context'' for simplicity. The assessment focuses on the context's pertinence to a given image $v$ and sentence $s$, alongside its applicability in MSA tasks. We employ the LLM-based metric, \ie \textbf{LLM-as-a-Judge}~\cite{vicuna2023} to quantify the quality of \textit{context}. Specifically, we craft a prompt for GPT-4V~\cite{openai2023gpt4} to compare our context with that provided by RAG and PKG. The detailed experimental settings can be found in Appendix~\ref{sec:rag_experiment_setting}. As shown in Fig.~\ref{fig:rag_ana}, our context ``Our Context'' significantly beats the RAG and PKG counterparts, demonstrating its superiority. We provide the examples of context provided by different methods in Appendix~\ref{sec:context_case}. Besides analyzing the contexts' pertinence of different methods, we report their downstreaming performance on MSA tasks in Table~\ref{tab:exp_rag}. Clearly, the MSA performance with our context is the best. 
The results above illustrate that \textit{\textbf{our method can provide more precise context, thus bringing better MSA performance.}}

\paragraph{Analyzing Effects of Contextual Fusion.}
In Table~\ref{table:fusion_strategy}, we experimentally explore different fusion strategies, including $mean(P_i,\hat{P}_i)$ (``\textbf{Average}''), $max(P_i,\hat{P}_i)$ (``\textbf{Max}''), Jensen-Shannon divergence~\cite{menendez1997jensen} (``\textbf{JS}''), conditional cross-mutual information score $f_{cxmi}$~\cite{wang2023learning} (``\textbf{CXMI}''), and our Context Fusion (``\textbf{CF}''). 
For JS, we calculate the JS divergence of $P_i$ with the uniform distribution to serve as the fusion weight, \ie $\beta$. As for CXMI, if $f_{cxmi} > 1.1$\footnote{In preliminary study, we grid-searched values ranging from 0.5 to 2.0, and 1.1 performs best on the dev set, thus leaving as our default setting.}, we adopt $(y_i, P_i)$ as our ultimate prediction, otherwise, we use $(\hat{y}_i,\hat{P}_i)$ as the final prediction. The results show that \textit{\textbf{our Contextual Fusion module performs the best among all competitive alternatives, confirming its effectiveness.}}


\begin{table}[tp]
\centering
\begin{tabular}{lll}
\toprule
        \multirow{2}{*}{\bf Method}          & \multicolumn{2}{c}{\bf Twitter2015} \\ \cline{2-3}
                  & \multicolumn{1}{c}{\textbf{\textit{Acc.}}}            & \multicolumn{1}{c}{\textbf{\textit{F1}}}                      \\ \hline
\multicolumn{1}{l}{AoM}               & \multicolumn{1}{l}{80.00}          & 75.20           \\ \hdashline
\multicolumn{1}{l}{\quad -w/ Average}           & 80.33 {\small (\green{+0.33})}          & 70.69 {\small (\red{-4.51})}         \\
\multicolumn{1}{l}{\quad -w/ Max}               & 80.56 {\small (\green{+0.56})}          & 76.81 {\small (\green{+1.61})} \\
\multicolumn{1}{l}{\quad -w/ JS}                & 80.78 {\small (\green{+0.78})}         & 77.39  {\small (\green{+2.19})}            \\
\multicolumn{1}{l}{\quad -w/ CXMI}       &   79.66 {\small (\red{-0.34})}    &  75.74  {\small (\green{+0.54})} \\
\multicolumn{1}{l}{\quad -w/ \textbf{CF}} & \textbf{81.45} {\small (\green{+1.45})}         & \textbf{78.12} {\small (\green{+2.92})}             \\ \bottomrule \toprule
\multicolumn{1}{l}{MMICL}             & 75.98          & 72.71                \\ \hdashline
\multicolumn{1}{l}{\quad -w/ Average}           & 75.53 {\small (\red{-0.45})}         & 72.33  {\small (\red{-0.38})}           \\
\multicolumn{1}{l}{\quad -w/ Max}               & 75.42 {\small (\red{-0.56})}         & 72.16  {\small (\red{-0.55})}          \\
\multicolumn{1}{l}{\quad -w/ JS}                & 77.09 {\small (\green{+1.11})}         & 73.92  {\small (\green{+1.21})}         \\
\multicolumn{1}{l}{\quad -w/ CXMI}       &   76.31 {\small (\green{+0.33})}    &  73.37  {\small (\green{+0.66})} \\
\multicolumn{1}{l}{\quad -w/ \textbf{CF}} & \textbf{77.32}  {\small (\green{+1.34})}        & \textbf{74.18} {\small (\green{+1.47})}            \\ \bottomrule
\end{tabular}
\caption{\textbf{Ablation study of different fusion strategies.} ``JS'' denotes Jensen-Shannon divergence. ``CF'' denotes our Contextual Fusion. }
\vspace{-10pt}
\label{table:fusion_strategy}
\end{table}

\subsection{When and Why Does Our Method Work?}
\label{subsec:analysis}
To better understand when and why our method works? we conduct extensive analysis to provide the following insights:

\begin{figure}[t]
    \centering
    \includegraphics[width=0.9\linewidth]{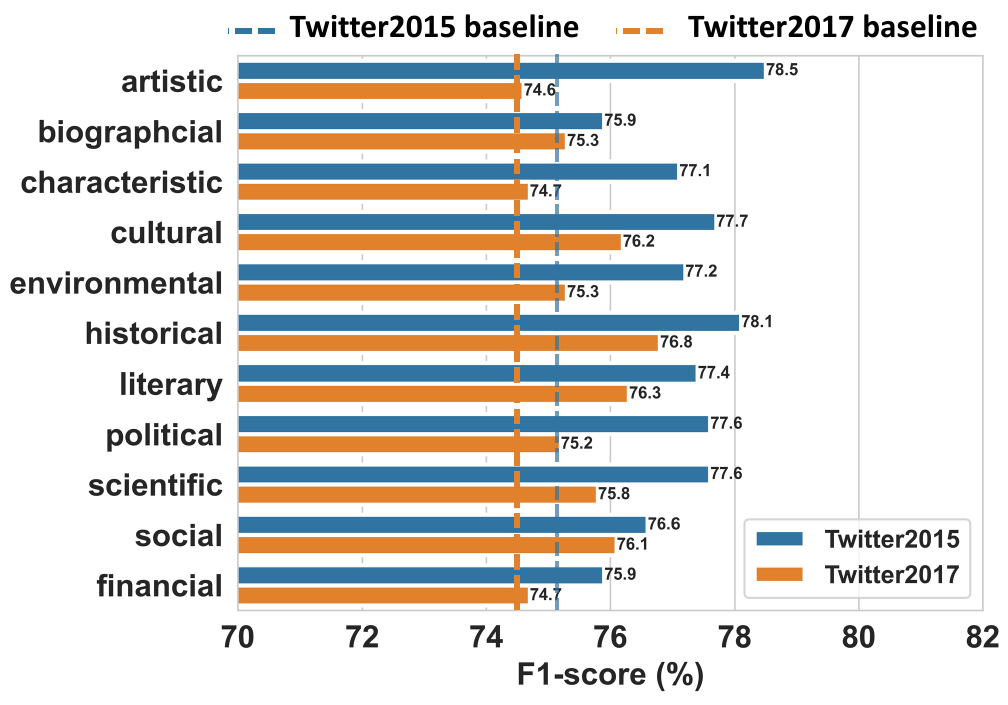}
    \caption{\textbf{Effects of different types of world knowledge.} We analyse the effect of different types of world knowledge by applying WisdoM to AoM. The orange dash line and blue dash line represent the F1-score of vanilla AoM on Twitter2015\&2017 respectively.}
    \label{fig:twitter_world_knowledge}
\end{figure}
\begin{figure}[t]
    \centering
    \includegraphics[width=0.86\linewidth]{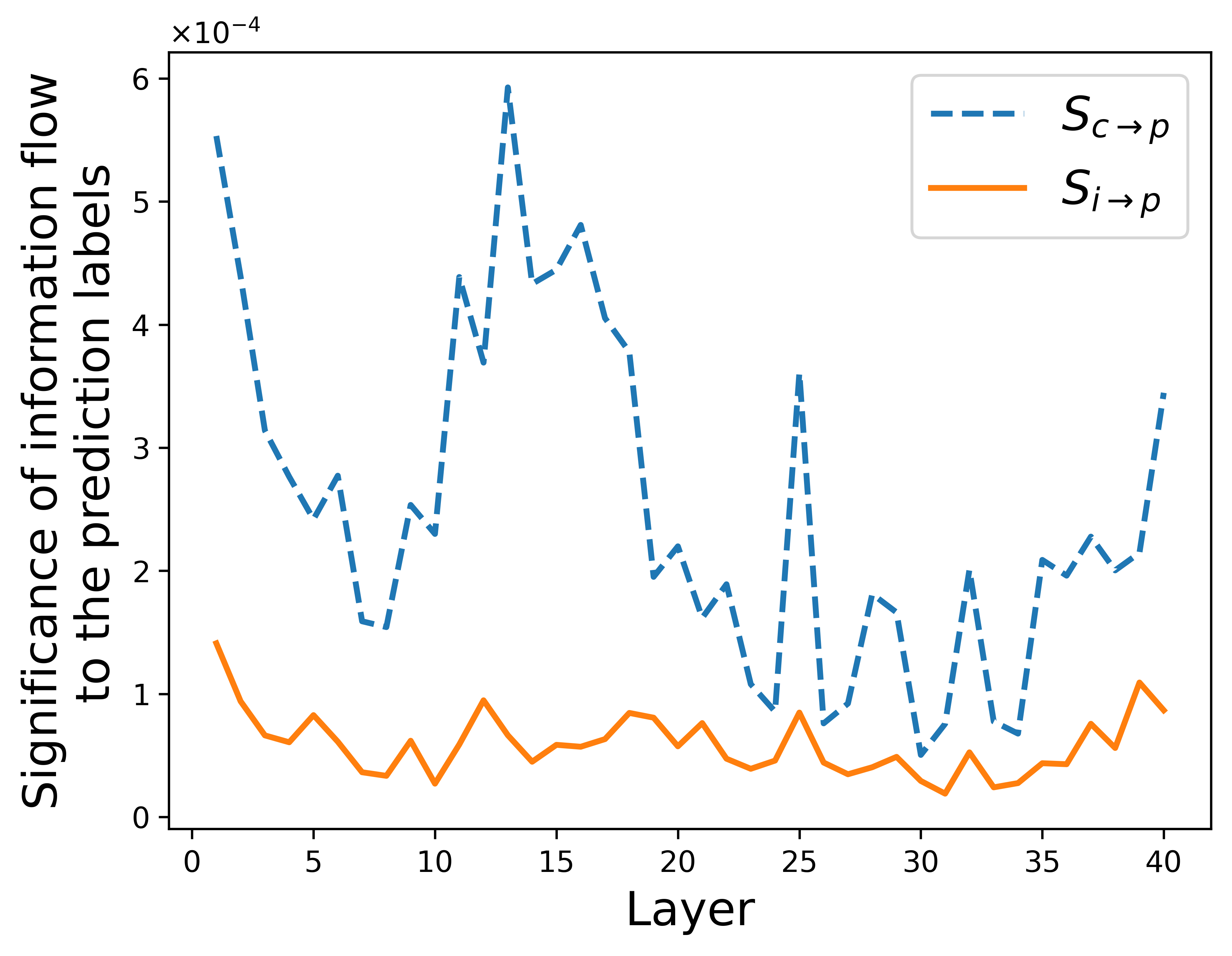}
    \caption{\textbf{Comparison of context ($S_{c \rightarrow p}$) and input's ($S_{i \rightarrow p}$) correlation to the final prediction across layers} in LLaVA-v1.5 on Twitter2015. High score means a strong correlation with final decision-making.}
    \label{fig:saliency_score_twitter2015}
\end{figure} 
\begin{figure}[t]
    \centering
    \includegraphics[width=1.\linewidth]{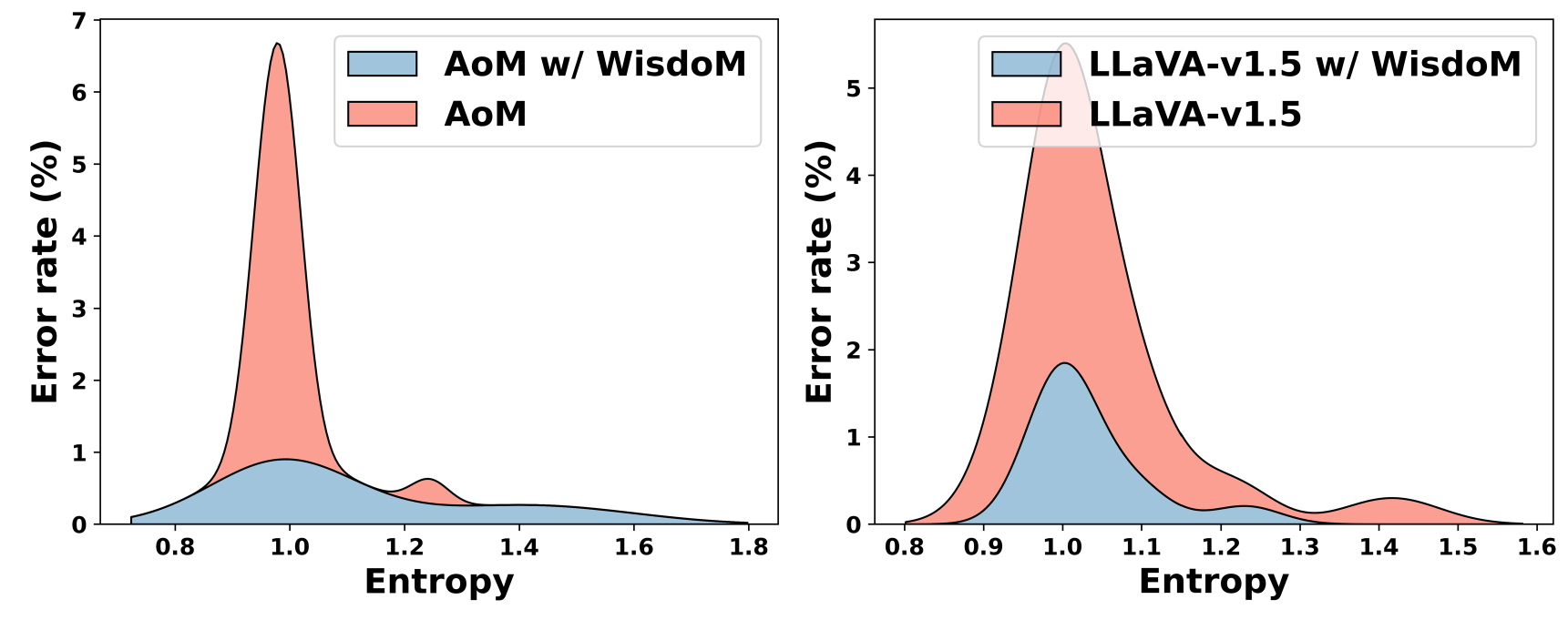}
    \caption{\textbf{Visualizing of error rate for hard samples ($\delta\leq 0.3$)} on Twitter2015 benchmark. }
    \label{fig:error_analysis}
\end{figure}

\paragraph{\textit{World Knowledge Enhances MSA, while Domain-related Knowledge is more Helpful.}}
We explore the effects of different types of world knowledge from 11 perspectives (artistic, biographical, etc.), upon AoM. Fig.~\ref{fig:twitter_world_knowledge} shows that 1) nearly all types of extra knowledge enhance the MSA performance, 2) historical knowledge significantly enhances MSA on Twitter datasets. We conjecture that this is because tweets often convey sentiment through historical references, and domain-related world knowledge is more helpful. To verify our hypothesis, we conduct experiments on Financial PhraseBank (FPB,~\citealp[]{malo2014good}) dataset and Twitter financial sentiment validation dataset~\cite{twitter2022finance}, and find that financial and historical knowledge are the two types of world knowledge with the greatest gain (bringing 4.2\% and 2.1\% improvements on average F1 scores, respectively), confirming our conjunction. Details can be refer to Appendix~\ref{subsection:financial_analysis}.

\paragraph{\textit{Context is Dominant for Prediction.}}
To draw a clearer picture of the information flow for MSA, we calculate $S_{c\rightarrow p}$ and $S_{i\rightarrow p}$ to represent the mean significance of information flow from \textit{context} (\textit{c}) and original input (\textit{i}) to the prediction labels (\textit{p}) respectively (\citet{simonyan2014deep,wang-etal-2023-label}, detailed in Appendix~\ref{subsection:s_score}). Fig.~\ref{fig:saliency_score_twitter2015} reveals that the significance of the information flow from \textit{context} to the prediction label is remarkably high than its \textit{input} counterpart, suggesting that \textit{context} outweighs image and sentence when making the final prediction. Results of other datasets show identical trends and can be found in Appendix~\ref{subsection:S_on_twitter17_msed}.

\paragraph{\textit{WisdoM Effectively Reduces the Uncertainty of Hard Samples.}}
To further explore how our WisdoM affects the hard samples, we visualize the error rate within high entropy in Fig.~\ref{fig:error_analysis}. After integrating WisdoM, the error rate is significantly decreased compared with the baseline (AoM and LLaVA-v1.5), demonstrating that our WisdoM effectively reduces the uncertainty of hard samples and improves performance.
\section{Conclusion}
In this paper, we propose a simple and effective plug-in framework \wisdom to enhance the ability of MSA. \wisdom follows three stages: Prompt Templates Generation, Context Generation, and Contextual Fusion. Firstly, we employ ChatGPT to generate prompt templates, enabling the large vision-language model to produce pertinent contextual world knowledge (referred to as \textit{context}) derived from both the image and sentence. Subsequently, we incorporate this \textit{context} using Contextual Fusion, minimizing the introduction of noise in the process. We empirically demonstrated the effectiveness and universality of the WisdoM on a series of widely used benchmarks. 

\section*{Limitation}
Our work has several potential limitations. Although our experiments revealed the enhanced performance of the multimodal sentiment analysis model with the introduction of context, the adaptive incorporation of context requires further exploration. Additionally, there is a need for further research into introducing models capable of handling additional modalities (\eg speech).

\section*{Ethics Statement}
We take ethical considerations very seriously. This paper focuses on improving multimodal sentiment analysis by making the most of large vision-language models. All experiments are conducted on open datasets and the findings and conclusions of this paper are reported accurately and objectively. Thus, we believe that this research will not pose ethical issues.


\normalem
\bibliography{arxiv_version}

\clearpage

\appendix
\section{Implementation Details}
\subsection{Details of Datasets}
In this work, we conduct experiments covering aspect-level (Twitter2015 and Twitter2017) and sentence-level (MSED) MSA benchmarks. We present the statistics of all datasets in Table~\ref{table:dataset_statics}. Then, each dataset is described as:
\paragraph{Twitter2015 \& 2017.}
Twitter2015 and Twitter2017 datasets encompass multimodal tweets, wherein each tweet comprises textual content, an accompanying image, embedded aspects, and corresponding sentiment annotations for each aspect. These aspects are categorized into labels from the set \{negative, neutral, positive\}.
\paragraph{MSED.}
MSED comprises 9190 pairs of text and images sourced from diverse social media platforms, including but not limited to Twitter, Getty Images, and Flickr.
\begin{table*}
\centering
    \begin{tabular}{cccccccccc}
    \toprule
     &  \multicolumn{3}{c}{\textbf{Twitter2015}} & \multicolumn{3}{c}{\textbf{Twitter2017}} & \multicolumn{3}{c}{\textbf{MSED}} \\ \cmidrule(lr){2-4} \cmidrule(lr){5-7} \cmidrule(lr){8-10} & \textit{\textbf{\#Train}} & \textit{\textbf{\#Dev}} & \textit{\textbf{\#Test}} & \textit{\textbf{\#Train}} & \textit{\textbf{\#Dev}} & \textit{\textbf{\#Test}} & \textit{\textbf{\#Train}} & \textit{\textbf{\#Dev}} & \textit{\textbf{\#Test}} \\ \hline
       \textit{\textbf{Negative}}  & 368 & 149 & 113 & 416 & 144 & 168 & 1939 & 308 & 613 \\
       \textit{\textbf{Neutral}}  & 1883 & 679 & 607 & 1638 & 517 & 573 & 1664 & 294 & 569 \\
       \textit{\textbf{Positive}} & 928 & 303 & 317 & 1508 & 515 & 493 & 2524 & 419 & 860  \\
       \textit{\textbf{Total}} & 3179 & 1122 & 1037 & 3562 & 1176 & 1234 & 6127 & 1021 & 2042 \\
       \bottomrule
    \end{tabular}
    \caption{Dataset statistics.}
    \label{table:dataset_statics}
\end{table*}
\subsection{Model Details}
To verify the effectiveness of our WisdoM, we apply it within two standard frameworks for modeling the MSA tasks. For task-specific framework, we conduct experiment using the AoM (105M,~\citealp[]{zhou-etal-2023-aom}) and ALMT (112.5M,~\citealp[]{zhang2023learning}). For general-purpose framework, we experiment on mPLUG-Owl2 (8.2B,~\citealp[]{ye2023mplugowl2}), LLaVA-v1.5 (13B,~\citealp[]{liu2023improvedllava}) and MMICL (14B,~\citealp[]{zhao2023mmicl}). The detailed model information is listed in Table~\ref{table:model_details}.
\begin{table*}
    \centering
    \begin{tabular}{ccc}
    \toprule
       Model  & Model Type & Source  \\ \hline
        AoM &  Task-specific & \href{https://github.com/SilyRab/AoM}{https://github.com/SilyRab/AoM} \\
        ALMT & Task-specific & \href{https://github.com/Haoyu-ha/ALMT}{https://github.com/Haoyu-ha/ALMT} \\
        LLaVA-v1.5 & General-purpose & \href{https://huggingface.co/liuhaotian/llava-v1.5-13b}{https://huggingface.co/liuhaotian/llava-v1.5-13b} \\
        mPLUG-Owl2 &  General-purpose & \href{https://huggingface.co/MAGAer13/mplug-owl2-llama2-7b}{https://huggingface.co/MAGAer13/mplug-owl2-llama2-7b} \\
        MMICL &  General-purpose & \href{https://huggingface.co/BleachNick/MMICL-Instructblip-T5-xxl}{https://huggingface.co/BleachNick/MMICL-Instructblip-T5-xxl} \\
    \bottomrule
    \end{tabular}
    \caption{Information of all models used in this study.}
    \label{table:model_details}
\end{table*}
\subsection{Training Details}
For a fair comparison, we fine-tune models introduced in \S~\ref{sec:models} on target datasets. We optimize our model with AdamW~\cite{loshchilov2018decoupled}. The learning rate is grid searched in \{1e-5, 2e-5, 7.5e-5, 1e-4\}, and batch size is in \{16, 32, 256\}. It should be noted that \textit{contexts} are not incorporated during the training stage.
\subsection{Inference Details}
For task-specific methods~\cite{zhou-etal-2023-aom,zhang2023learning}, we employ the same inference procedures as their original works.
For general-purpose methods~\cite{liu2023improvedllava,zhao2023mmicl,ye2023mplugowl2}, we compute the likelihood that an LVLM generates the content of this choice given the question. We select the choice with the highest likelihood as the model's prediction.
\subsection{Prompt Templates of World Knowledge}
\label{app:prompt_templates}
In Table~\ref{table:prompt}, we list the prompt templates which are provided by ChatGPT in prompt templates generation stage. In the context generation stage, we replace ``[x]'' with sentence $s_i$ and a special image token (\eg ``<image>'') is inserted at the beginning of the prompt template. 
\subsection{Pseudo Code for Contextual Fusion}
Algorithm~\ref{algo:contextual_fusion} provides the pseudo-code of Contextual Fusion. The simplicity of our method requires only a few lines of code.
\begin{algorithm}[hbt]
\scriptsize
\SetAlCapNameFnt{\footnotesize}
\SetAlCapFnt{\footnotesize}
\caption{Python-style pseudo-code for Contextual Fusion.}
\label{algo:contextual_fusion}
\SetAlgoLined
    \PyComment{p\_o: numpy array, represents $P_i$} \\
    \PyComment{p\_c: numpy array, represents $\hat{P}_i$} \\
    \PyComment{alpha: float, the threshold of choosing hard sample} \\
    \PyComment{beta: float, interpolation coefficient} \\
    \PyCode{def contextual\_fusion(p\_o, p\_c, alpha, beta):}\\
    \Indp   
         \PyComment{step1 : calculate delta} \\
         \PyCode{delta = 2 * np.max(p\_o) + np.min(p\_o) - 1} \\ 
         \PyCode{if delta > alpha:}\\
         \Indp
              \PyCode{return p\_o} \\
        \Indm
        \PyComment{step2 : calculate the final prediction} \\
         \PyCode{p\_f = p\_o + beta * (p\_c - p\_o)} \\ 
         \PyCode{return p\_f} \\  
    \Indm 
\end{algorithm}

\begin{table*}[ht]
    \centering
    \begin{tabularx}{\linewidth}{cX}
        \toprule
        Type name & Prompt template \\ \hline
        \multirow{1}{*}[-3.0ex]{Artistic} & Identify and discuss any artistic movements or styles that influenced the creation of the image. Explore how the artist's choice of style aligns with or deviates from prevalent artistic trends of the time. Sentence: [x]. \\ \hline
        \multirow{1}{*}[-3.0ex]{Biographcial} & Delve into the backgrounds of individuals associated with the image and text. Explore the biographies of artists, authors, or other relevant figures, and discuss how their life experiences shaped the creation and interpretation of the work. Sentence: [x] \\ \hline
        \multirow{1}{*}[-3.0ex]{Character} & Focus on characters within the image and sente. Analyze their personalities, relationships, and potential character development. Discuss how the visual and textual elements contribute to character portrayal. Sentence: [x] \\ \hline
        \multirow{1}{*}[-3.0ex]{Cultural} & Explore how the image and sentence reflect or represent aspects of a particular culture. Discuss the cultural significance, traditions, or values implied by the elements in the image and sentence. Sentence: [x] \\ \hline
        \multirow{1}{*}[-3.0ex]{Environmental} & Examine the environmental elements within the image and sentence, discussing ecological factors, environmental changes, or the relationship between human activities and the depicted setting. Sentence: [x] \\ \hline
        \multirow{1}{*}[-1.5ex]{Historical} & Give you an image and sentence, you can provide historical context, important events, and relevant background information related to the image and sentence. Sentence: [x] \\ \hline
        \multirow{1}{*}[-3.0ex]{Literary} & Conduct a literary analysis of the sentence, exploring themes, symbolism, and narrative techniques. Discuss how the words complement or contrast with the visual elements in the image. Sentence: [x] \\ \hline
        \multirow{1}{*}[-3.0ex]{Political} & Examine the political during the time the image and text were created. Discuss any political events, movements, or ideologies that may have influenced the content and tone of the work. Sentence: [x] \\ \hline
        \multirow{1}{*}[-3.0ex]{Scientific} & Investigate the scientific elements within the image, delving into discoveries, advancements, or breakthroughs related to the subject matter mentioned in the sentence. Sentence: [x] \\ \hline
        \multirow{1}{*}[-3.0ex]{Social} & Investigate the image and text as a form of social commentary. Analyze how the work reflects or critiques social issues, norms, or inequalities prevalent at the time of creation. Sentence: [x] \\ \hline
        \multirow{1}{*}[-1.5ex]{Financial} & Give you a sentence and image, you should provide related financial knowledge. Sentence: [x] \\
        \bottomrule
    \end{tabularx}
    \caption{\textbf{Example of the prompt template} generated in stage 1 and used in stage 2.}
    \label{table:prompt}
\end{table*}

\subsection{RAG Experiment Setup}
\label{sec:rag_experiment_setting}
In \S~\ref{sec:effect_context}, we compare our WisdoM with two RAG-based methods: (1) a naive RAG~\cite{lewis2020retrieval}, which initially searches for relevant documents related to a given question and then employs a generator to predict an answer; (2) PKG~\cite{luo2023augmented}, an advanced RAG method that incorporates a knowledge-guided module, allowing for information retrieval without modifying the parameters of language models. The experimental setting is described as below.

\paragraph{Knowledge Sources.} 
The source of knowledge for our experiments is the Wikipedia-Image-Text (WIT) dataset~\cite{10.1145/3404835.3463257}, which has published in 2021. This dataset comprises images from Wikipedia, along with their alt-text captions and contextualized text passages. 
\paragraph{Methods.}
For naive RAG, we use off-the-shelf Contriever-MSMARCO~\cite{izacard2022unsupervised} as the textual retriever and CLIP-ViT as the visual retriever. Specifically, we utilize CLIP-ViT and Contriever-MSMARCO for encoding query $q$ and  knowledge source, and employ Maximum Inner Product Search (MIPS,~\citealp[]{guo2020accelerating}) to find the five nearest neighbors (\textit{knowledge}) to $q$ within the knowledge source. For PKG, we use LLaVA-v1.5 (13B) as knowledge-guided module fine-tuning on WIT and then generate the \textit{knowledge} according to the image $v$ and sentence $s$. Subsequently, we directly predict sentiment polarity by incorporating \textit{knowledge}.
\paragraph{Evaluation Setting.}
To evaluate the relevance of the context to a specific image $v$ and sentence $s$, we employ LLM-based metric, \ie LLM-as-a-Judge~\cite{vicuna2023}. Specifically, a prompt is crafted for GPT-4V~\cite{openai2023gpt4} to assess the winning rates of our context in comparison to those derived from RAG-based methods. The detailed prompt can be found in Table~\ref{table:evaluation_prompt}. We also evaluate performance on MSA tasks.
\begin{table}[thb]
\begin{tabularx}{\linewidth}{X}
\toprule
\textbf{Evaluation Prompt}     \\ \hline
**System**: In this task, you will be asked to compare the relevance of two paragraphs to determine which one is more pertinent to the provided source sentence and image and benefits the sentiment analysis task the most. There are three options for you to choose from: \\
1. Context1 is better. If you think Context 1 is more relevant to the source sentence and image and benefits the sentiment analysis task.\\
2. Context2 is better. If you think Context 2 is more relevant to the source sentence and image and benefits the sentiment analysis task.\\
3. Context1, Context2 are the same: If you think Context1, Context2 have the same relevance to the source sentence and image, then choose this option. \\
\\
**Your answer is a JSON DICT that has one key: answer. For example: \{"answer": "x. Context x is better."\}**\\
\\
**INPUT**\\
Source Sentence: ``[s]''\\
\\
Context1: ``[x1]''\\
\\
Context2: ``[x2]''\\
\\
**OUTPUT** \\
\bottomrule
\end{tabularx}%
\caption{\textbf{The Evaluation Prompt} we used for GPT-4V. [s] represents the input sentence. [x1] and [x2] represent the context generated by different methods.}
\label{table:evaluation_prompt}
\end{table}

\subsection{Calculation of $S_{c \rightarrow p}$ and $S_{i \rightarrow p}$}
\label{subsection:s_score}
To measure the significance of \textit{context} and original input (\ie image and sentence), 
we use $S_{c \rightarrow p}$ and $S_{i \rightarrow p}$ for highlighting critical token interactions. Following previous work~\cite{simonyan2014deep,wang-etal-2023-label}, we use the Taylor expansion~\cite{michel2019sixteen} to calculate the score for each element of the attention matrix:
\begin{equation}
    I_l = \sum_{h} |A_{h,l}^\mathsf{T}\frac{\partial \mathcal{L}(x)}{\partial A_{h,l}}|.
\end{equation}
Here, $A_{h,l}$ represents the value of the attention matrix of the $h$-th attention head in the $l$-th layer, $x$ is the input, and $\mathcal{L}(x)$ is the loss function. We calculate the saliency matrix $I_l$ for the $l$-th layer by averaging across all attention heads. $I_l(k,j)$ denotes the importance of the information flow from the $j$-th word to the $k$-th word. We propose two quantitative metrics based on $I_l$. The definitions of the two quantitative metrics are below.

\noindent \textbf{$S_{c \rightarrow p}$, the mean significance of information flow from \textit{context}($c$) to the prediction label ($p$).}
\begin{equation}
\begin{aligned}
    & S_{c \rightarrow p} = \frac{\sum_{(k,j)\in C_{cp}}I_{l}(k,j)}{|C_{cp}|},\\
    & C_{c,p} = \{(c,p): c \in \textit{context}\}.
\end{aligned}
\end{equation}

\noindent \textbf{$S_{i \rightarrow p}$, the mean significance of information flow from image ($v$) and sentence ($t$) to the prediction label ($p$).}
\begin{equation}
\begin{aligned}
    & S_{i \rightarrow p} = \frac{\sum_{(k,j)\in C_{ip}}I_{l}(k,j)}{|C_{ip}|},\\
    & C_{i,p} = \{(i,p): i \in [v, t]\}.
\end{aligned}    
\end{equation}
$S_{c \rightarrow p}$ and $S_{i \rightarrow p}$ indicate the intensity of information aggregation onto the prediction label. A high $S$ demonstrates strong information for final decision-making. 

\section{Additional Experimental Results}
\begin{figure}[t]
    \begin{center}
    \includegraphics[width=0.95\linewidth]{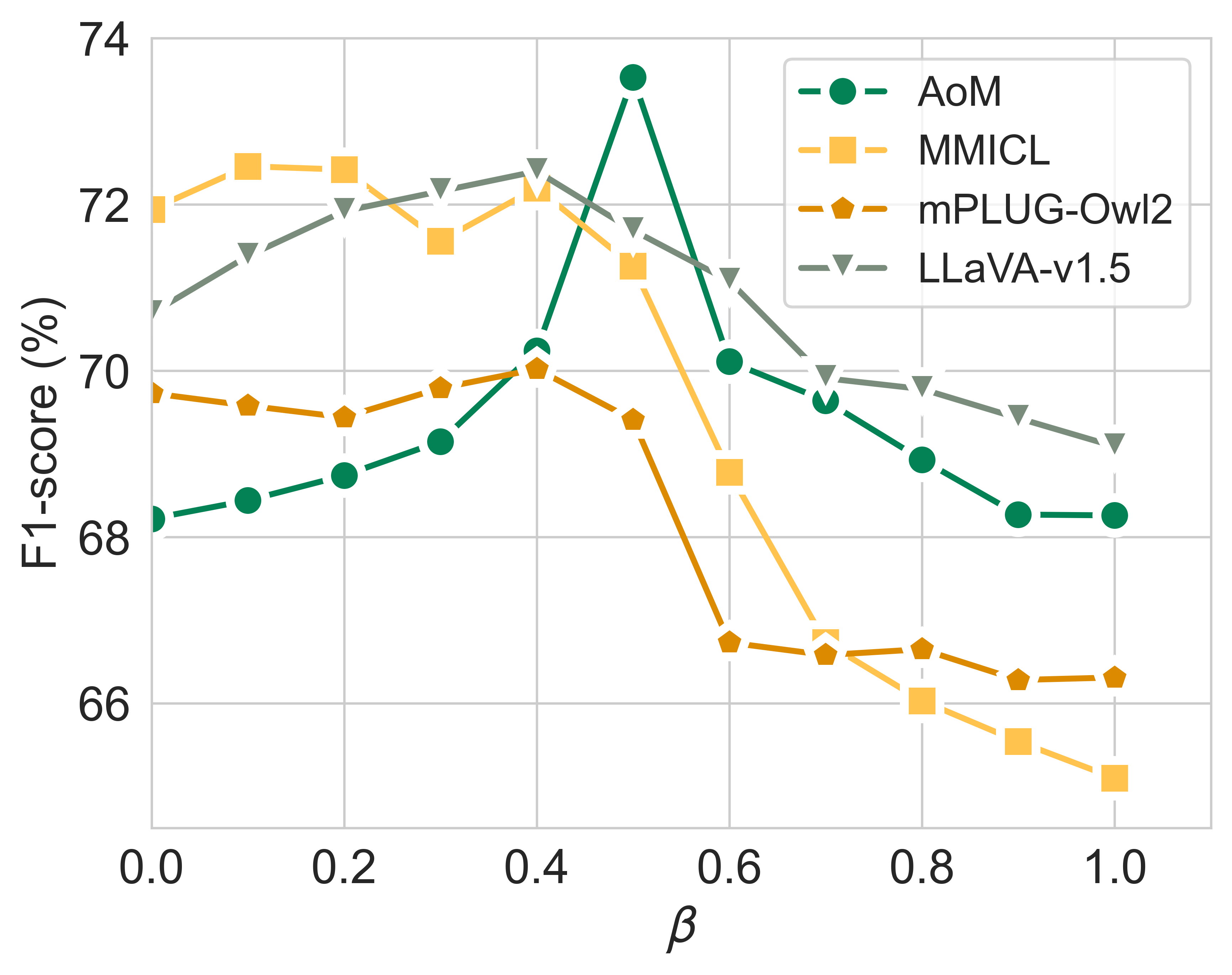}
    \end{center}
    \caption{
    \textbf{Effect of interpolation coefficient $\beta$.} We show how F1-score changes when varying the $\beta$ values.
    }
    \label{fig:beta_twitter}
\end{figure}
\begin{figure}[t]
    \centering
    \includegraphics[width=0.95\linewidth]{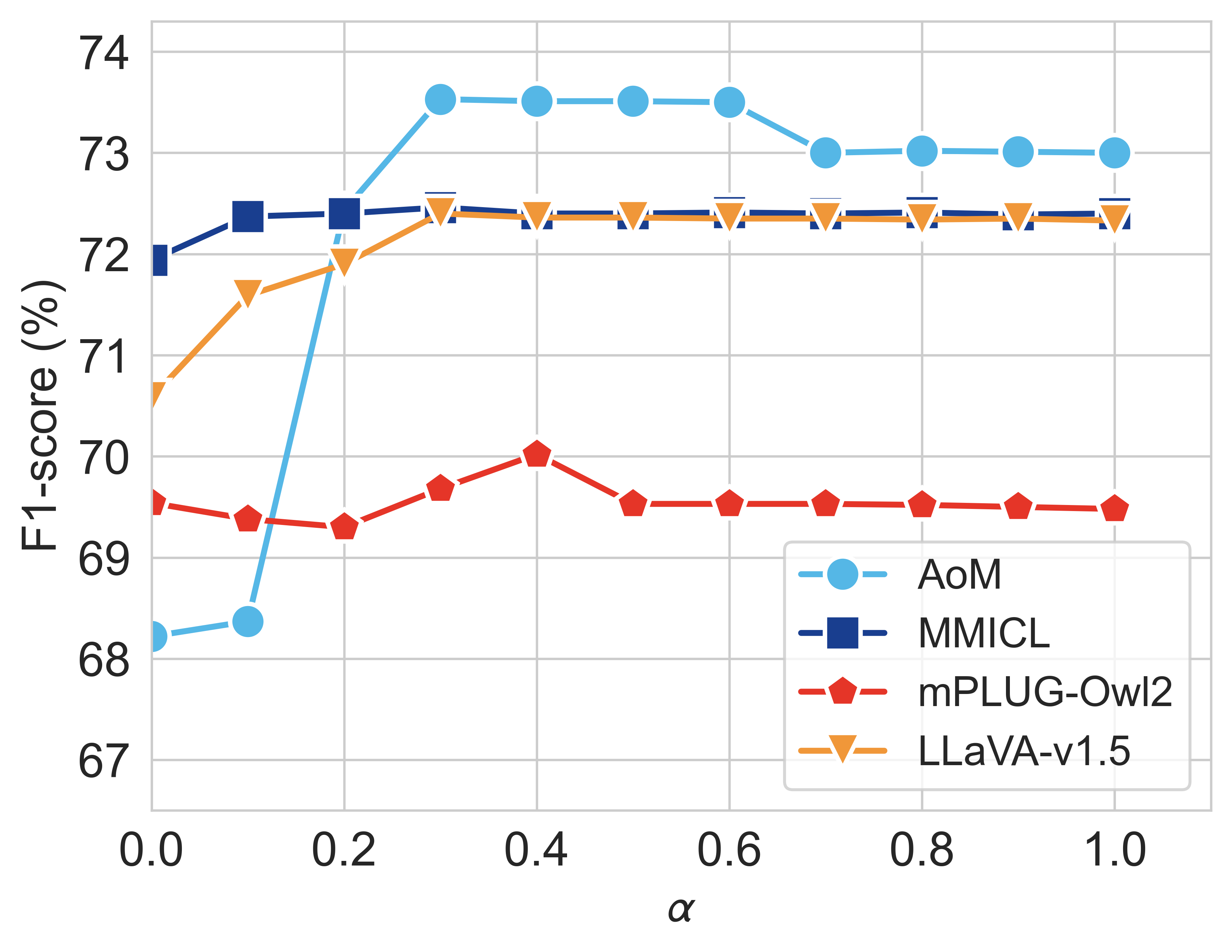}
    \caption{\textbf{Impact of uncertain threshold $\alpha$}, illustrating how the F1-score changes when varying $\alpha$.}
    \label{fig:alpha_twitter}
\end{figure}
\subsection{Hyperparameter Selection}
\label{section:hyperparameter}
In stage 3, Contextual fusion has two major hyperparameters interpolation coefficient $\beta$ and uncertain threshold $\alpha$. To systematically study the impact of $\beta$ and $\alpha$, we first fix $\alpha=0.3$ and search different configurations for the $\beta$. Then we fix $\beta$ to the optimal value in the Twitter2015 dev set and search for $\alpha$. 
\paragraph{Parameter Analysis of $\beta$.}
As shown in Fig.~\ref{fig:beta_twitter}, we find that: 1) different models are sensitive to the values of $\beta$. 2) $\beta$ values within the range of $[0.4,0.5]$ demonstrate strong performance among various models. 3) Excessive values of $\beta$ result in performance degradation. We conjecture that this decline may be caused by introducing excessive noise within the \textit{context}. For a fair comparison, we select the optimal value of $\beta$ on the dev set for evaluation.
\paragraph{Parameter Analysis of $\alpha$.}
Fig.~\ref{fig:alpha_twitter} shows that: 1) models exhibit insensitivity to the value of $\alpha$. 2) $\alpha$ values within the range of $[0.3,0.4]$ exhibit strong performance across diverse models. 3) There is no significant decrease in trends with increasing $\alpha$. Thus, we set $\alpha=0.3$ as our default setting.
\subsection{Scalability of WisdoM}
\begin{figure}[h]
    \centering
    \includegraphics[width=1.\linewidth]{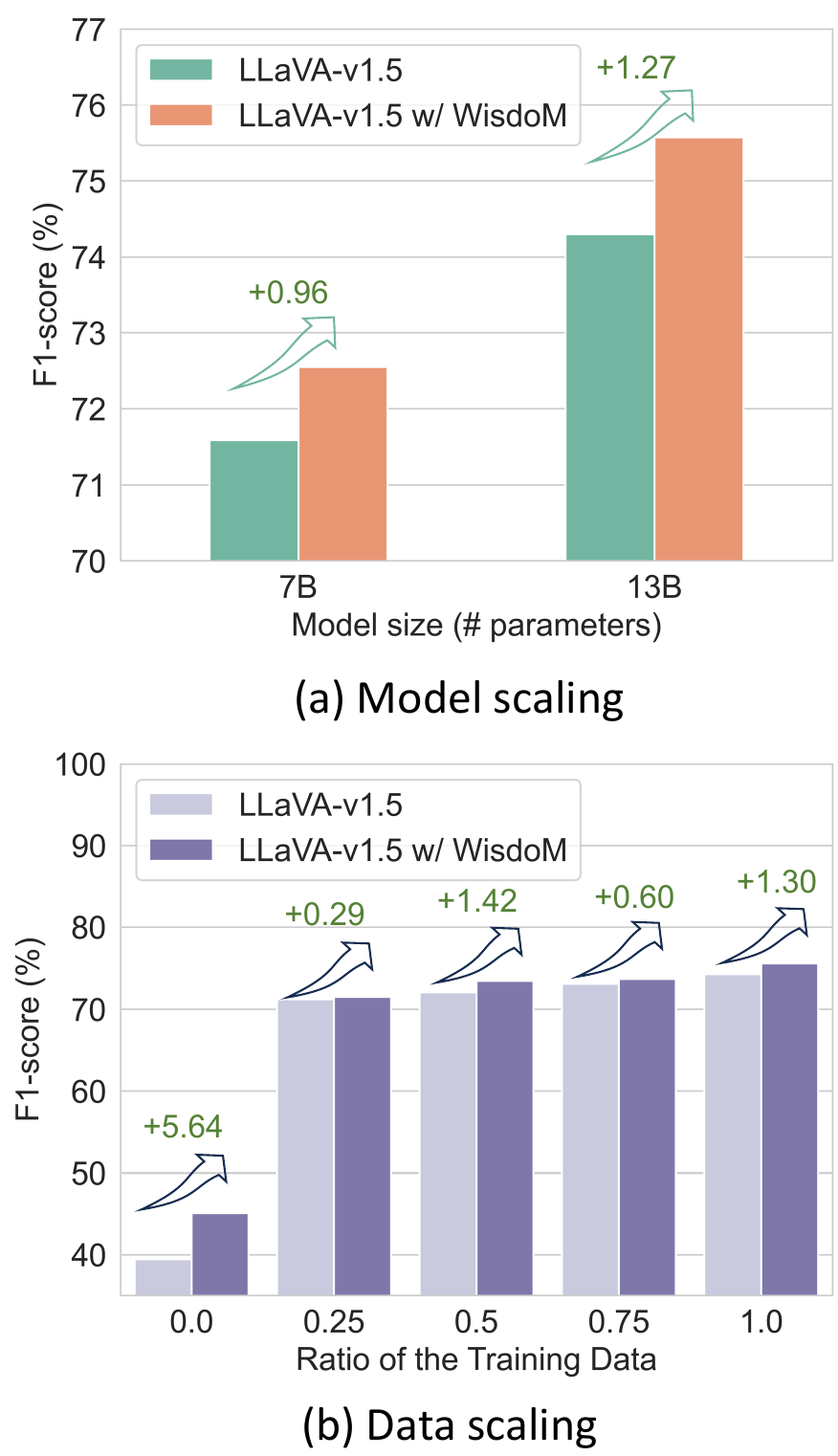}
    \caption{\textbf{Performance of scaling WisdoM} on Twitter2015 with different a) model and b) data scales.}
    \label{fig:scale}
\end{figure}

Our plug-in method is data- and model-agnostic, therefore, it is expected to be highly scalable. Here we scale our \wisdom up to different model sizes and data volumes.
\paragraph{Performance on Different Model Sizes.}
We experiment with scaling the model size to see if there are ramifications when operating at a larger scale. Fig.~\ref{fig:scale} (a) reveals that the performance increases as the LVLM size increases. In addition, we found that as the size of the model increased, the performance gains became more pronounced.
\paragraph{Performance on different Data Volumes.}
We conduct experiments on different ratios of training data to verify the robustness of WisdoM. As shown in Fig.~\ref{fig:scale} (b), we find that even when only 25\% of the training data was used, our method WisdoM resulted in a 0.29\% improvement.

\subsection{Exploring Contexts Derived from Various LVLMs} 
To explore the relationship between \textit{context} and LVLMs capability, we conduct experiments on AoM using \textit{contexts} derived from mPLUG-Owl2 (8.2B) and LLaVA-v1.5 (13B). As depicted in the Table~\ref{table:model_generate_context}, the results show that \textbf{the stronger the capability of LVLMs, the more accurate and helpful the generated context is for MSA.}
\begin{table}[htb]
    \begin{tabular}{lcccc}
        \toprule
            \multirow{2}{*}{Method} & \multicolumn{2}{c}{\textbf{Twitter2015}} & \multicolumn{2}{c}{\textbf{Twitter2017}} \\ \cmidrule(lr){2-3} \cmidrule(lr){4-5} & \textit{\textbf{Acc.}} & \textit{\textbf{F1}} & \textit{\textbf{Acc.}} & \textit{\textbf{F1}} \\ \hline
            mPLUG-Owl2  & 76.8  & 72.3 &  74.2  & 73.0 \\
            LLaVA-v1.5  & 77.9 & 74.3  & 74.6 & 74.3 \\ \hline
            AoM  & 80.0  & 75.2  & 75.9  & 74.5 \\ \hdashline
            \quad -w/ Context$_{m}$  &  81.2  &  77.8  & 76.4  & 75.2 \\
            \quad -w/ Context$_{L}$   &  \textbf{81.5}  & \textbf{78.1}   & \textbf{77.6}   & \textbf{76.8} \\
        \bottomrule
    \end{tabular}
    \caption{\textbf{Comparison of contexts derived from different LVLMs.} ``Context$_{m}$'' represents the context derived from mPlUG-Owl2. ``Context$_{L}$'' represents the context derived from LLaVA-v1.5.}
    \label{table:model_generate_context}
\end{table}

\subsection{Training with Context}
\begin{figure}[hbt]
    \centering
    \includegraphics[width=1.\linewidth]{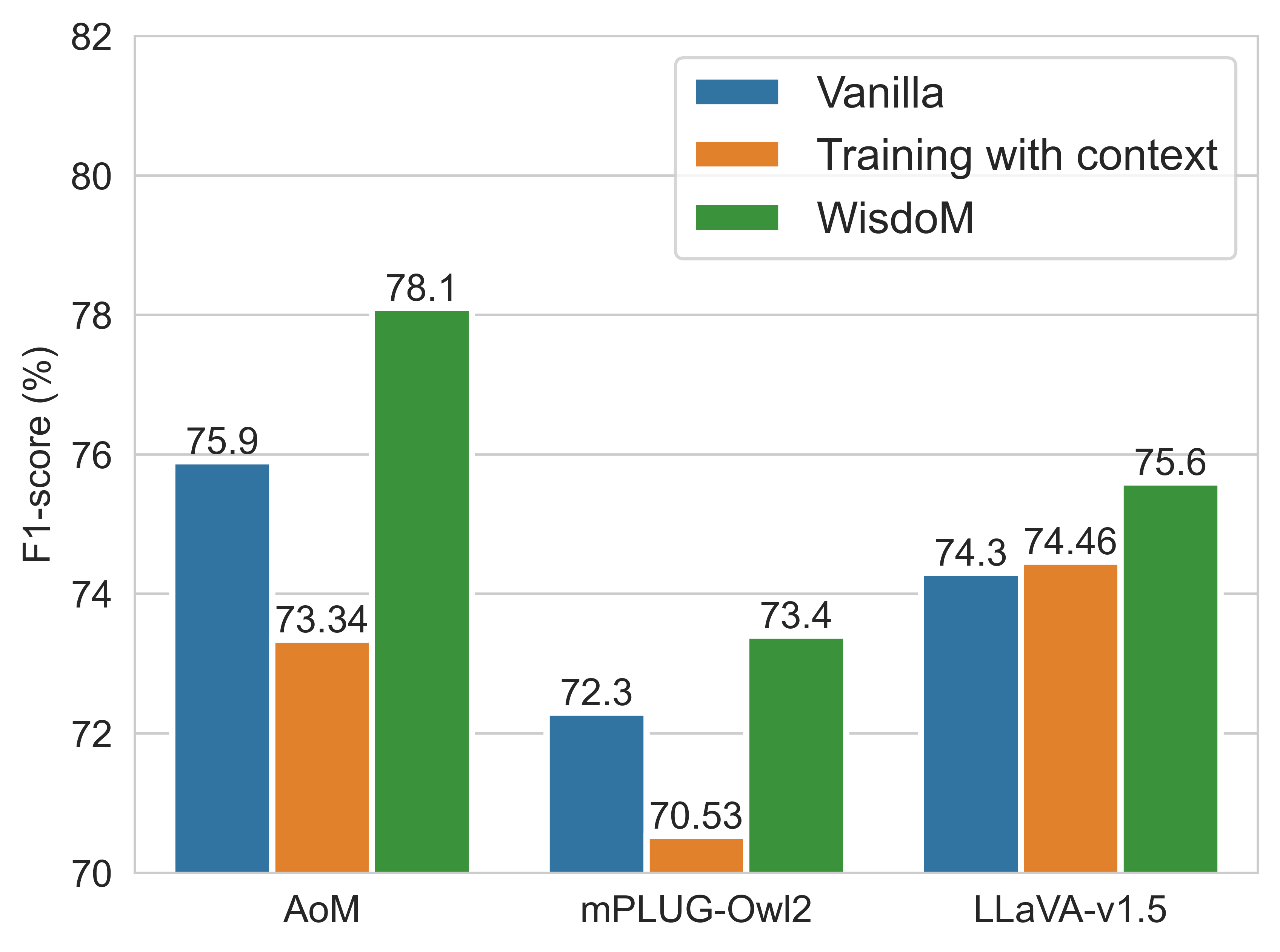}
    \caption{\textbf{Comparison of WisdoM with vanilla and training with context on Twitter2015.} We utilize the historical world knowledge generated by LLaVA-v1.5 for training and observed significant decreases in AoM and mPLUG-Owl2.}
    \label{fig:train_with_context}
\end{figure}

To investigate the impact of context incorporation during training, we conduct experiments with three different setups: 1) vanilla, which without incorporation of context during training and inference; 2) training with context, which incorporation of context into training; 3) our WisdoM, to leverage world knowledge during inference phrase. The results, depicted in Fig.~\ref{fig:train_with_context}, demonstrate that our WisdoM consistently outperforms the vanilla across all models. However, the performance of training with context, specifically AoM and mPLUG-Owl2, experienced a significant decline. The possible reason is that the noisy nature of the context used during training. We conjecture that developing a method to effectively filter out this noise could potentially ameliorate the performance of models trained with context.

\subsection{Additional World Knowledge Analysis}
\label{subsection:financial_analysis}
In \S~\ref{subsec:analysis}, we analyzed the effect of different types of world knowledge in MSA tasks and conjecture that domain-related world knowledge is more helpful. To verify our hypothesis, we perform experiment on financial sentiment analysis. The experimental setting is described as below.
\paragraph{Datasets.}
Following~\cite{zhang2023enhancing}, we perform instruction tuning on Twitter Financial News dataset~\cite{twitter2022finance} and FiQA~\cite{fiqa}. We conduct evaluation on Financial PhraseBank dataset (FPB,~\citealp[]{malo2014good}) and Twitter financial news sentiment validation dataset (Twitter Val,~\citealp[]{twitter2022finance}). Twitter Financial News Sentiment Dataset focuses on financial sector tweets, while FiQA consists of 961 annotated samples. The Financial PhraseBank dataset includes 4840 randomly selected samples from financial news articles in the LexisNexis database.
\paragraph{Baselines.}
The experiment includes three different types of baselines: 1) Direct generation without knowledge, without providing any knowledge for a given task and ask the MSA model to response directly; 2) Generation with retrieval financial knowledge, retrieveing related knowledge from external knowledge sources (\eg \textit{News Source}, \textit{Research Publication Platforms} and \textit{Social Media Platforms}), following the approach of prior works~\cite{zhang2023enhancing}; 3) our WisdoM with non-financial \textit{context}, employing gpt-3.5-turbo to generate historical knowledge in our stage 2.
\paragraph{Implementation Details.}
Since FPB and Twitter Val are textual sentiment analysis benchmarks, we adopt the same experimental setup as~\cite{zhang2023enhancing}, using Llama-7B~\cite{touvron2023llama} as our MSA model. In our WisdoM, we employ gpt-3.5-turbo to generate historical and financial knowledge respectively and select the hyperparameter on FiQA validation dataset. The performance metric is F1 score.
\paragraph{Result.}
Table~\ref{table:financial_benchmark} shows that: 1) our WisdoM consistently brings improvement across all benchmarks among various types of knowledge, 2) financial knowledge brings more improvement than historical knowledge. The results prove that \textbf{\textit{domain-related knowledge is more helpful.}}
\begin{table}[thbp]
\centering
\resizebox{1.\linewidth}{!}{
\begin{tabular}{lcc}
\toprule
Method & \textbf{FPB} & \textbf{Twitter Val} \\ \hline 
Llama-7B                       &     72.1        &     78.7      \\ \hdashline
\quad -w/ RAG                &  75.0   &  81.2   \\
\quad -w/ WisdoM{\small (Historical)}                &      74.8      &  80.2  \\
\quad -w/ \textbf{WisdoM{\small (Financial)}}                                &      \textbf{76.0}       &  \textbf{83.1}  \\
\bottomrule
\end{tabular}
}
\caption{\textbf{Experimental results of Llama-7B with different methods on FPB and Twitter Val.} History and Financial represent the incorporation of history and financial knowledge respectively.}
\label{table:financial_benchmark}
\end{table}

\subsection{$S_{c \rightarrow p}$ and $S_{i \rightarrow p}$ of Other datasets}
\label{subsection:S_on_twitter17_msed}
\begin{figure}[h]
    \centering
    \includegraphics[width=1.\linewidth]{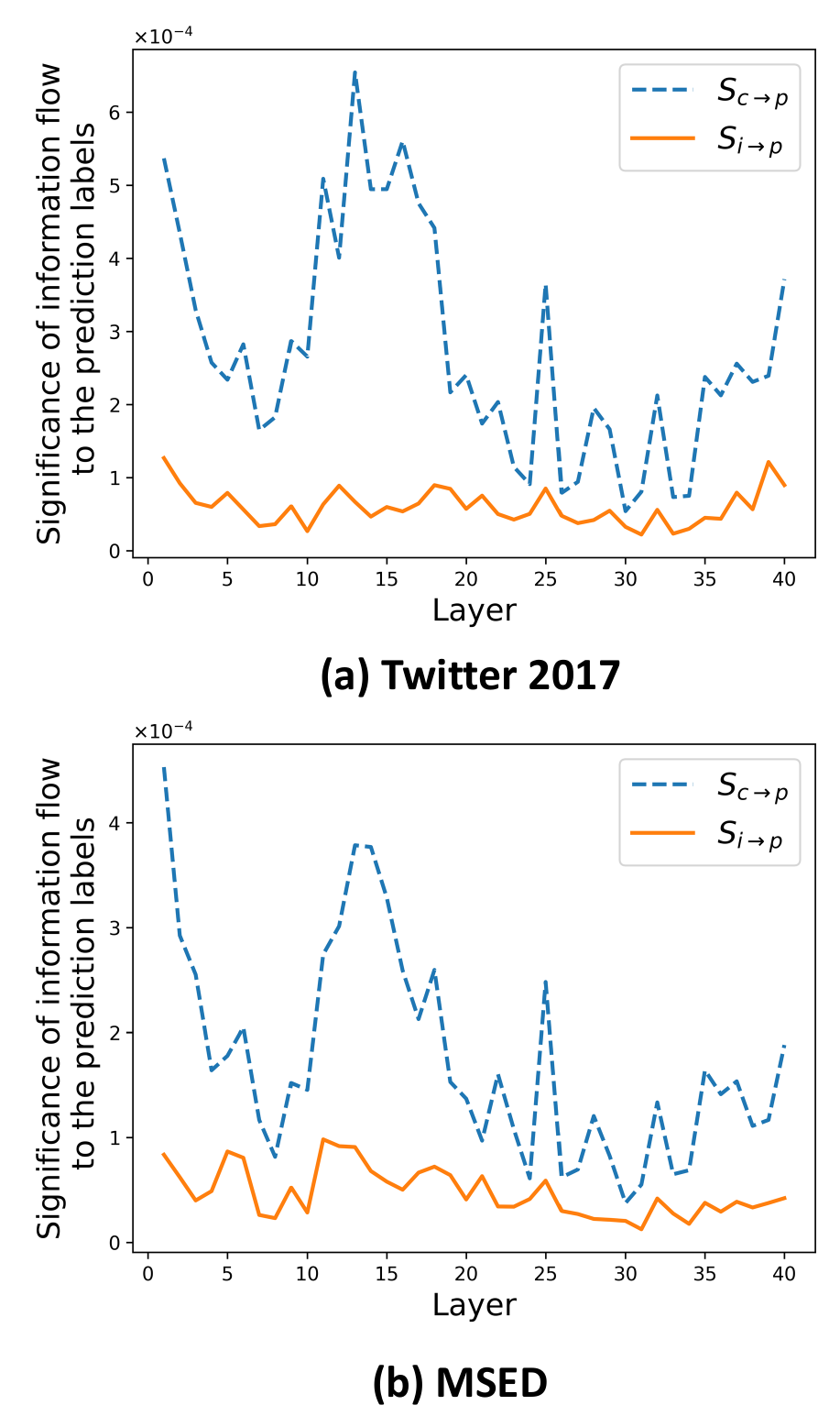}
    \caption{\textbf{$S_{c \rightarrow p}$ and $S_{i \rightarrow p}$ on Twitter2017 and MSED.} $S_{c \rightarrow p}$ represents the information flow from \textit{context} to the prediction label and $S_{i \rightarrow p}$ represents information flow from image and sentence to prediction label.}
    \label{fig:attn_weight_twitter17_msed}
\end{figure}

Fig.~\ref{fig:attn_weight_twitter17_msed} illustrates the $S_{c \rightarrow p}$ and $S_{i \rightarrow p}$ on Twitter2017 and MSED. $S_{c \rightarrow p}$ is prominent, while $S_{i \rightarrow p}$ is less significant.
\section{Case Study}

\subsection{Example of Contexts}
\label{sec:context_case}
Examples of context from RAG-based methods and our WisdoM are presented in Table~\ref{table:case_context}. It is evident that our WisdoM offers more precise context, providing detailed descriptions of image elements. In contrast, RAG-based methods exhibit a lack of specificity in image details and demonstrate weak relevance to the associated sentences.
\subsection{Qualitative Examples of Aspect-Level MSA}
\label{section:case_study}
We present two qualitative examples from Table~\ref{table:case_study_aspect} showcasing historical knowledge. In the first, adding background information about Aleppo aided in accurate decision-making. Conversely, in the second, the inclusion of satirical content led to misclassification. However, with Contextual Fusion, LLaVA-v1.5 ultimately make the correct decision.
\subsection{Qualitative Examples of Sentence-Level MSA}
Table~\ref{table:msed_case_study} illustrates two instances of incorporating scientific world knowledge into the MSED dataset. In the first example, ALMT initially predicts negative, but incorporating information on masks' positive role in epidemic prevention leads to a positive sentiment prediction. In the second example, excessive focus on coffee benefits causes misclassification by LLaVA-v1.5. However, with Contextual Fusion, LLaVA-v1.5 ultimately predicts correctly.
\subsection{Qualitative Examples of Financial Sentiment Analysis}
In addition, we also provide examples in Financial Sentiment Analysis in Table~\ref{table:twitterfinancial_case_study}. It can be observed that for samples with higher expertise, Llama-7B tends to predict incorrectly, but by introducing explanations for these samples, Llama-7B can predict correctly.
\begin{table*}[]
    \centering
    \begin{tabular}{p{0.1\linewidth}|p{0.85\linewidth}}
    \toprule
      Sentence   & RT @ shanilpanara : Bus Selfie on the way to Harry Potter Studios @ WFCTrust @ NCSEast \# ShareYourSummer \\ \hline
      Image    & \includegraphics[width=0.3\linewidth]{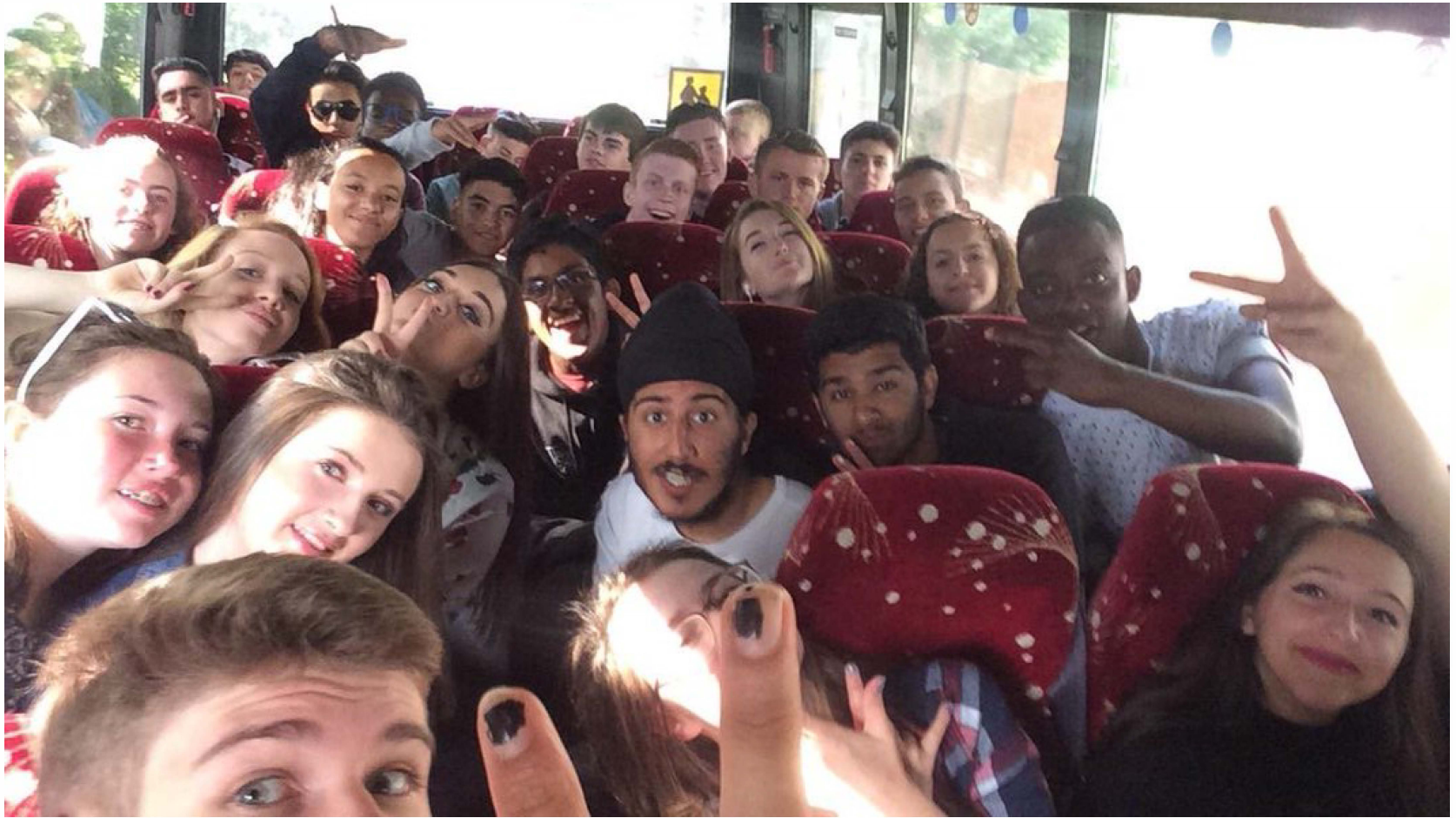}  \\ \hline
       RAG  & Harry Potter and the Forbidden Journey is a motion-based dark ride located in The Wizarding World of Harry Potter-themed areas of Islands of Adventure in Orlando, Florida; Universal Studios Hollywood in Universal City, California, Universal Studios Japan in Osaka, Japan and is upcoming to Universal Studios Beijing in Beijing, China. The ride takes guests through scenes and environments in and around Hogwarts Castle from the Harry Potter series of books and films. Mark Woodbury, president of Universal Creative, described the ride as an in-depth look at the world of Harry Potter, which utilizes never-before-seen technology which transforms ``the theme park experience as you know it''. The ride first opened at Islands of Adventure with The Wizarding World of Harry Potter on June 18, 2010; subsequent versions opened at Universal Studios Japan on July 15, 2014, and at Universal Studios Hollywood on April 7, 2016. \\ \hline
       PKG & A picture of a person with a camera and a big lens in front of them, taken by a tourist. \\ \hline
       WisdoM & The image shows a large group of people, including men and women, sitting on a bus. They are all wearing black and white clothing, which suggests that the photo might have been taken during a specific time period or event when black and white clothing was more common. The group appears to be diverse, with people of different ages and ethnicities, which indicates that the bus might be a public transportation vehicle or a special event bus. In the context of historical events, the black and white clothing could imply that the photo was taken during the early to mid-20th century when colour photography was not widely available. The group of people might be attending a social event, a gathering, or a public transportation service during that time. The presence of a diverse group of people on the bus also reflects the social dynamics and cultural diversity of the period. \\
    \bottomrule
    \toprule
    Sentence & RT @ wemssecnirp : Now I really am like Cleopatra \# hypnotizing \# findom \# goddess \\ \hline
    Image  &  \includegraphics[width=0.3\linewidth]{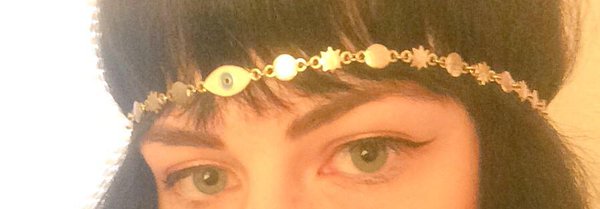} \\ \hline
    RAG   &  Cleopatra has been the subject of literature, films, plays, television programs, and art. Only those with Wikipedia articles are cited. \\ \hline
    PKG  &  A portrait of Cleopatra, the Queen of Egypt, wearing a golden headdress and a necklace. The portrait was drawn in the style of Raphael and engraved on copper. \\ \hline
    WisdoM  &  The image shows a woman wearing a gold headband with a large, ornate design featuring eyes. This headband is likely inspired by ancient Egyptian or African culture, as it resembles the traditional headpieces worn by women in these regions. The eyes on the headband may symbolize protection, wisdom, or the ability to see beyond the physical realm. In ancient Egypt, headbands were often worn by both men and women as a symbol of status, power, and religious beliefs. They were made from various materials, such as gold, silver, or precious stones, and often featured intricate designs and symbols. Similarly, in African cultures, headbands have been used as a form of adornment and expression of identity, with designs and patterns often representing specific tribes, clans, or regions. The woman in the image is wearing a modern interpretation of these traditional headbands, which demonstrates the enduring influence of ancient cultures on contemporary fashion and design. \\ 
    \bottomrule
    \end{tabular}
    \caption{\textbf{Examples of contexts generated by RAG, PKG and our WisdoM.} Our contexts provide a detailed description that matches the elements present in images.}
    \label{table:case_context}
\end{table*}

\begin{table*}
    \centering
    \begin{tabular}{p{0.25\linewidth}|p{0.7\linewidth}}
    \toprule
         Sentence & RT @ AHedengren : \# \underline{Aleppo} before and after . \# Syria . \\ \hline
         Image   & \includegraphics[width=0.3\linewidth]{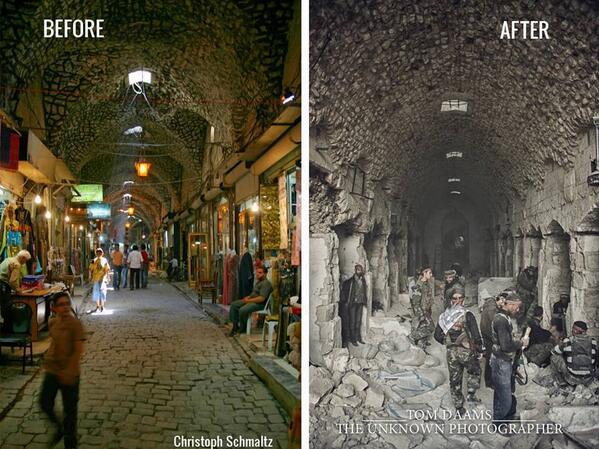} \\ \hline
         Aspect  &   Aleppo \\ \hline
         \textit{Context} & \green{The image shows a comparison of the same street in Aleppo, Syria, before and after the devastating war.} The before picture depicts a bustling and lively street with people walking and cars driving, while the after picture displays a desolate and destroyed street with rubble and ruins. The stark contrast between the two images highlights the devastating impact of the war on the city and its residents. \green{Aleppo, once a thriving and historic city, has been severely affected by the ongoing Syrian Civil War. The conflict has led to the destruction of infrastructure, displacement of residents, and loss of life.} The image serves as a powerful reminder of the human cost of war and the need for peaceful resolution to conflicts. \\ \hline
         Label  & Negative \\ \hline
         AoM   &   \red{Neutral} \\ \hdashline
         \quad + \textit{Context} & \green{Negative} \\ 
         \quad + Contextual Fusion   & \green{Negative} \\ 
    \bottomrule
    \toprule
        Sentence & \underline{Donald Trump} will win in a landslide. * The cartoonist behind ‘ Dilbert ’ explains why. \\ \hline
         Image   & \includegraphics[width=0.3\linewidth]{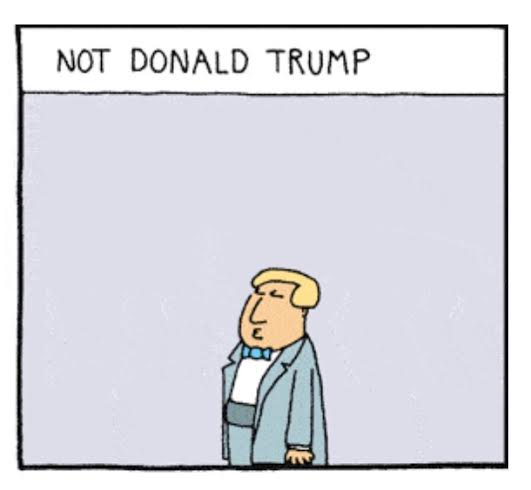} \\ \hline
         Aspect  &   Donald Trump \\ \hline
         \textit{Context} & The image features a cartoon of a man in a suit and tie, standing in front of a wall with a sign that says "Not Donald Trump." \red{This cartoon is likely a political satire or a commentary on the 2016 United States presidential election, as Donald Trump was one of the main candidates running for office at that time.} The cartoon could be a representation of the political climate, opinions, or reactions towards Donald Trump and his campaign. It might also be a way to express the artist's or the creator's perspective on the election, the political landscape, or the candidates involved. \\ \hline
         Label  & Positive \\ \hline
         LLaVA-v1.5   &   \green{Positive} \\ \hdashline
         \quad + \textit{Context} & \red{Negative} \\ 
         \quad + Contextual Fusion   & \green{Positive} \\ 
         \bottomrule
    \end{tabular}
    \caption{\textbf{Qualitative examples of aspect-level MSA}. Two cases with predictions by SOTA methods and our method on Twitter2015 and Twitter2017. We initially integrate only the \textit{context} and then introduce Contextual Fusion. Clues to answering the input are highlighted in \green{green} within the \textit{Context}. The \textit{Context} that will disrupt MSA is highlighted in \red{red}.}
     \label{table:case_study_aspect}
\end{table*}

\begin{table*}[thb]
    \centering
    \begin{tabular}{p{0.25\linewidth}|p{0.7\linewidth}}
    \toprule
      Sentence   & Group of teenagers posing showing their protective face masks during COVID-19 coronavirus epidemic spread. \\ \hline
       Image  & \includegraphics[width=0.3\linewidth]{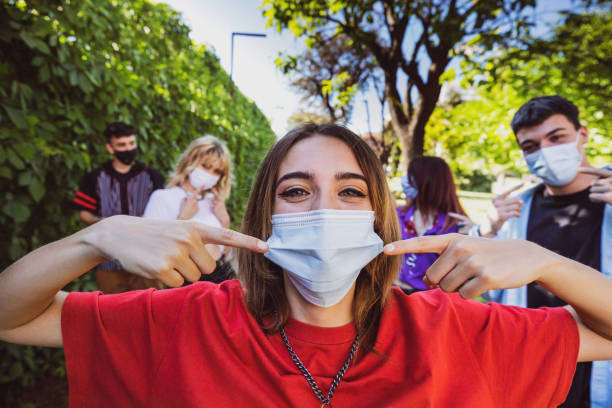} \\ \hline
       \textit{Context} & The image features a group of teenagers posing with their protective face masks, which are an essential part of personal protective equipment (PPE) during the COVID-19 pandemic. \green{The use of face masks has been a crucial measure to reduce the transmission of the virus, as they help prevent respiratory droplets from spreading into the air. Scientifically, the effectiveness of face masks in preventing the spread of COVID-19 has been extensively studied. Research has shown that wearing a face mask can significantly reduce the transmission of respiratory viruses, including COVID-19. The masks work by filtering out respiratory droplets that may contain the virus, preventing them from reaching other people's noses and mouths.} In conclusion, the image of the group of teenagers posing with their face masks highlights the importance of wearing PPE during the COVID-19 pandemic. The scientific evidence supporting the effectiveness of face masks in preventing the spread of respiratory viruses, including COVID-19, has been instrumental in promoting their use as a crucial public health measure. \\ \hline
       Label & Positive \\ \hline
       ALMT & \red{Negative} \\ \hdashline
       \quad + \textit{Context} & \green{Positive} \\ 
       \quad + Contextual Fusion  & \green{Positive} \\
    \bottomrule
    \toprule
    Sentence & Handsome young man standing in the kitchen with a cup of coffee at home. \\ \hline
    Image & \includegraphics[width=0.3\linewidth]{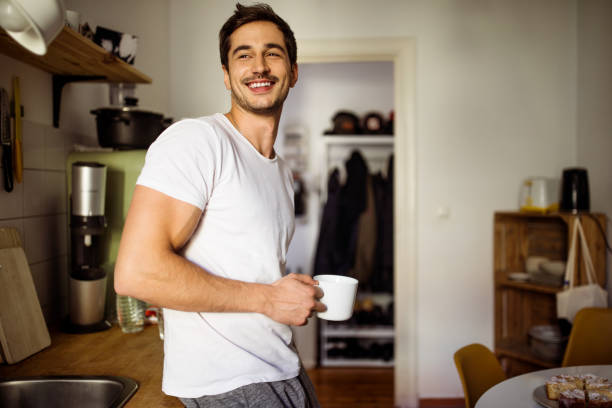} \\ \hline
    \textit{Context} & The image features a handsome young man standing in a kitchen with a cup of coffee. \red{Recent studies have shown that moderate coffee consumption can have positive effects on cognitive function, memory, and mood. Additionally, coffee has been found to have potential health benefits, such as reducing the risk of certain diseases, including type 2 diabetes and certain types of cancer. In terms of advancements in coffee production and consumption, there has been a growing interest in sustainable and eco-friendly coffee practices.} This includes the use of organic farming methods, fair trade practices, and the development of new technologies to reduce waste and improve the efficiency of coffee production. \\ \hline
    Label & Neutral \\ \hline
    LLaVA-v1.5 & \green{Neutral} \\ \hdashline
    \quad + \textit{Context} & \red{Positive} \\
    \quad + Contextual Fusion & \green{Neutral} \\
    \bottomrule
    \end{tabular}
    \caption{\textbf{Qualitative examples of MSED.} We initially integrate only the \textit{context} and then introduce Contextual Fusion. Clues to answering the input are highlighted in \green{green} within the \textit{Context}. \red{Red} represents incorrect prediction.}
    \label{table:msed_case_study}
\end{table*}
\begin{table*}[bht]
    \centering
    \begin{tabular}{p{0.25\linewidth}|p{0.7\linewidth}}
    \toprule
    Sentence & Last year the company raised its turnover to approximately 7 million litas EUR 2 mln, from 6.1 million litas in 2004. \\ \hline
    \textit{Context} & The text you provided discusses the company's turnover, which refers to the total amount of money generated from sales of goods or services. It's important to note that turnover is different from profit, as it represents the company's total sales before deducting any expenses. In this case, the company's turnover increased from approximately 6.1 million litas in 2004 to 7 million litas (equivalent to EUR 2 million) last year. This indicates a positive trend in the company's sales performance. \green{Increasing turnover can be a sign of business growth and improved market demand for the company's products or services.} \\ \hline
    Label & Positive \\ \hline
    Llama-7B & \red{Neutral} \\ \hdashline
    \quad + \textit{Context} & \green{Positive} \\
    \quad + Contextual Fusion & \green{Positive} \\
    \bottomrule
    \toprule
      Sentence   & \$XOM (+5.8\% pre) Exxon cuts full-year capex forecast by 30\%, maintains long-term outlook - SA \\ \hline
       \textit{Context} & Exxon Mobil Corporation (XOM) has increased by 5.8\% in pre-market trading. The company has announced a 30\% reduction in its full-year capital expenditure forecast while maintaining its long-term outlook. This move reflects the company's response to changing market conditions and its commitment to long-term sustainability. Capital expenditure forecasts are important indicators of a company's investment plans and financial health. Investors often monitor these forecasts closely to assess a company's growth prospects and financial management. \green{The market's positive reaction to this news suggests that investors may view Exxon's decision as a prudent and forward-thinking strategy.} \\ \hline
       Label & Positive \\ \hline
       Llama-7B & \red{Negative} \\ \hdashline
       \quad + \textit{Context} & \green{Positive} \\ 
       \quad + Contextual Fusion & \green{Positive} \\
    \bottomrule
    \toprule
    Sentence & Stock Market Update: Stock market drifts in record territory \\ \hline
    \textit{Context} & The stock market drifting in record territory typically indicates a period of stability and potential growth. \green{Investors may interpret this as a positive sign for the economy and the companies listed on the stock exchange.} However, it's important to consider various factors such as interest rates, inflation, and geopolitical events that could impact market movements. Investors should also diversify their portfolios and stay informed about market trends to make well-informed decisions. It's advisable to consult with a financial advisor for personalized guidance based on individual financial goals and risk tolerance. \\ \hline
    Label & Positive \\ \hline
    Llama-7B & \red{Negative} \\ \hdashline
    \quad + \textit{Context} & \green{Positive} \\
    \quad + Contextual Fusion & \green{Positive} \\
    \bottomrule
    \end{tabular}
    \caption{\textbf{Qualitative examples of Financial PhraseBank (FPB) and Twitter financial news sentiment validation (Twitter Val).} Clues to answering the input are highlighted in \green{green} within the \textit{Context}. \red{Red} represents incorrect prediction.}
    \label{table:twitterfinancial_case_study}
\end{table*}

\end{document}